\documentclass[conference]{IEEEtran}
\usepackage{cite}
\usepackage{amsmath,amssymb,amsfonts}
\usepackage{graphicx}
\usepackage{textcomp}
\usepackage{xcolor}
\usepackage{fancyhdr}
\usepackage[hyphens]{url}
\usepackage{algorithm}
\usepackage[noend]{algpseudocode}
\usepackage{subcaption}
\usepackage{comment}

\usepackage{hyperref}
\hypersetup{
    colorlinks=true,
    linkcolor=blue,
    filecolor=magenta, 
    urlcolor=cyan,
}

\def\BibTeX{{\rm B\kern-.05em{\sc i\kern-.025em b}\kern-.08em
    T\kern-.1667em\lower.7ex\hbox{E}\kern-.125emX}}

\begin{document}

\title{Privacy-Preserving Inference in Machine Learning Services Using Trusted Execution Environments}

\author{\IEEEauthorblockN{Krishna Giri Narra}
\IEEEauthorblockA{\textit{Electrical and Computer Engineering} \\
\textit{University Of Southern California}\\
LA, CA, USA\\
narra@usc.edu\\}
\and
\IEEEauthorblockN{Zhifeng Lin}
\IEEEauthorblockA{\textit{Electrical and Computer Engineering} \\
\textit{University Of Southern California}\\
LA, CA, USA \\}
\and
\IEEEauthorblockN{Yongqin Wang}
\IEEEauthorblockA{\textit{Electrical and Computer Engineering} \\
\textit{University Of Southern California}\\
LA, CA, USA \\}
\and
\IEEEauthorblockN{Keshav Balasubramaniam}
\IEEEauthorblockA{\textit{Electrical and Computer Engineering} \\
\textit{University Of Southern California}\\
LA, CA, USA \\}
\and
\IEEEauthorblockN{Murali Annavaram}
\IEEEauthorblockA{\textit{Electrical and Computer Engineering} \\
\textit{University Of Southern California}\\
LA, CA, USA \\}
}


\maketitle

\pagestyle{plain}

\begin{abstract}
Preserving the privacy of user input sent to a cloud-based machine learning inference service is a critical need. One approach for private inference is to run the trained model within a secure hardware enclave. The user then sends encrypted data into the enclave, which is then decrypted within the enclave before running the inference entirely within the enclave. 
Secure enclaves, like Intel SGX, however, impose several restrictions. First, enclaves can only access limited memory without relying on expensive paging, thereby limiting the size of the model than can be run efficiently.  Second, the increasing use of accelerators like GPUs and TPUs for inference will be curtailed in this mode of execution as accelerators currently do not provide enclaves.

To tackle these challenges, this work presents Origami, which provides privacy-preserving inference for large deep neural network (DNN) models through a combination of enclave execution, cryptographic blinding, interspersed with accelerator-based computation.  Origami partitions the ML model into multiple partitions. The first partition receives the encrypted user input within an SGX enclave. The enclave decrypts the input and then applies cryptographic blinding to the input data and the model parameters. Cryptographic blinding is a technique that adds noise to obfuscate data.  Origami sends the obfuscated data for computation to an untrusted GPU/CPU. The blinding and de-blinding factors are kept private by the SGX enclave, thereby preventing any adversary from denoising the data, when the computation is offloaded to a GPU/CPU. The computed output is returned to the enclave, which decodes the computation on noisy data using the unblinding factors privately stored within SGX. This process may be repeated for each DNN layer, as has been done in prior work Slalom. 

However, the overhead of blinding and unblinding the data is a limiting factor to scalability. Origami relies on the empirical observation that the feature maps after the first several layers can not be used, even by a powerful conditional GAN adversary to reconstruct input. Hence, Origami dynamically switches to executing the rest of the DNN layers directly on an accelerator without needing any further cryptographic blinding intervention to preserve privacy. We empirically demonstrate that using Origami, a conditional GAN adversary, even with an unlimited inference budget, cannot reconstruct the input. We implement and demonstrate the performance gains of Origami using the VGG-16 and VGG-19 models. Compared to running the entire VGG-19 model within SGX, Origami inference improves the performance of private inference from 11x while using Slalom to 15.1x.

\end{abstract}

\section{Introduction}
\label{sec:introduction}
Deep learning (DL) has made significant strides possible in computer vision, machine translation, robotics, healthcare, etc. Training on large volumes of data is necessary to make supervised deep learning models accurate. As a consequence, the development of DL models happens primarily at organizations that have access to large data sets. After training, the trained DL models may be deployed in the cloud to serve user requests. With the rise in MLaaS (Machine Learning as a Service) offerings by cloud vendors like Amazon AWS, Microsoft Azure and Google cloud, organizations may deploy pre-trained models in the cloud. Users need to send their data such as images and text for inference. While running in the cloud the DL models can be exposed to a wide attack surface consisting of malicious users, compromised hypervisors and physical snooping, leading to data leakage. 
Users expect the service providers to protect the confidentiality of their data. It is the responsibility of the service providers to meet the user expectations and not compromise privacy of user data accidentally or otherwise. Regulations like GDPR are attempting to enforce this requirement on all organizations handling private user data.

One approach to protect confidentiality of data is using cryptographic deep learning models \cite{miniONN, cryptoNets}. These DL models can process encrypted user data and as a result cannot leak confidential user data. The main limitation with this approach is that it can take orders of magnitude more processing time than non-cryptographic DL models, which is a severe limitation in user facing services.

Another technique to protect confidentiality is leveraging Trusted Execution Environments (TEE) like Intel SGX \cite{McKeen_intel_sgx}, ARM TrustZone \cite{Alves2004TrustZoneI} or Sanctum \cite{sanctum}. Classic techniques like data encryption can protect the data during it's storage and communication phases. TEEs complement these protections by protecting the data during computation phase. TEEs achieve this task by using a combination of hardware and software techniques to isolate and protect sensitive data and computations. These security features can be exploited to provide private inference capabilities. For instance, one approach for private inference is to run the complete trained model within an Intel SGX enclave, which is invisible to the cloud service provider or even a potential hacker with root access to the cloud server. The user then sends encrypted data into the enclave, which is then decrypted within the enclave before running the inference entirely within the enclave. 

However, running an entire model inference within a TEE provides several practical hurdles. First accelerators such as GPUs do not support trusted execution. Hence, applications cannot take advantage of the growing list of DL accelerators, such as GPUs and TPUs, if they rely purely on TEE to achieve privacy. Second, TEEs do not support efficient execution of arbitrarily large programs. Typically the program memory footprint is limited to a threshold. This threshold is less than 128 MB for Intel SGX enclaves. When the program memory limit exceeds the threshold then frequent swapping of data in and out of SGX leads to significant performance slowdowns. This overhead stems from the fact that moving pages in and out of SGX enclaves requires  decryption and encryption of data.

Popular DL models such as VGG-16 and VGG-19 \cite{VGG} have large memory footprint, larger than the SGX memory limit. The performance gap between running DL models within SGX compared to running DL model inference on an untrusted GPU/CPU is very high. As we demonstrate later, in our experimental setup running a VGG-19 model inference entirely within SGX is 105 times slower than running it on a GPU, and is 6.5 times slower than running it on a CPU. 
These drastic slowdowns makes it unpalatable to use SGX to run the entire inference models within SGX. 

\textbf{Model splitting:}
To minimize the negative performance impacts of SGX, hardware vendors recommend that a program be carefully split into sensitive and non-sensitive parts, and schedule only the sensitive part to be executed inside the TEE. They also recommend to minimize the sensitive code to reduce latency. Following this guidance specifically for deep neural network (DNN) models (a particular variant of DL models),  one approach we explored is to perform some of the DNN layers  within SGX and allow other layers to be executed outside of SGX. Since DNNs provide a clean layer abstraction it is possible to re-design the models such that only the first few layers are executed within a secure container while the remaining layers are executed in an untrusted container.
The computations that are performed outside of SGX can take advantage of accelerators such as GPUs. However, care must be taken such that the computations performed on a GPU, without the benefit of SGX protection, should still protect the privacy of user input. 

Given the above discussed limitations, in this paper, we propose \textit{Origami Inference}, a new inference framework that lowers the performance overhead of using TEE while protecting the privacy of the user data. It is built on Intel SGX, but comes with the flexibility to run models that are much larger than SGX physical protected memory, and can exploit unsecure accelerators, such as GPUs. Origami is built on the insight  that only the first few layers of the model contain most of the information that can be used to reconstruct the model's input and the output from deeper layers of the model can not help with input reconstruction. In Origami, a pre-trained model is split into two partitions. The first partition consists of multiple layers of a DNN. Each layer within the first partition  is split partially between GPU and SGX enclave. We use the Slalom approach~\cite{Tramr2018SlalomFV} to offload computationally intensive convolutions (basically matrix multiplications) on GPUs, while allowing the non-linear operations (such as ReLUs) to be performed within an enclave. Data privacy is preserved using Cryptographic blinding of the convolution data before offloading to the GPU (more details of blinding later). Unlike Slalom, which continues to split every DNN layer between GPU and SGX leading to unnecessary overheads, Origami dynamically switches to offloading the second partition of the DNN model to execute entirely on a GPU (or even a CPU). 

We present and evaluate a conditional GAN adversary model to verify if a user's input can be reconstructed from the intermediate data sent to the GPU. We demonstrate that it is possible to partition DNNs in such a way that the computation of the first partition of the model that runs using cryptographic blinding between GPU and SGX can be minimized. As an example, for VGG-16 model we find that it is sufficient to run the first 4 convolutional layers using blinding to protect input privacy. The remaining 9 convolutional layers and 3 fully connected layers can be executed completely in the open without input recovery capability. 

To summarize this paper makes the following contributions:
\begin{enumerate}
    \item We design and implement Origami framework that partitions DNN models between secure enclaves and unsecure accelerators.
    \item Origami uses input obfuscation to cooperatively execute between an enclave and untrusted CPU/GPU in the early layers of DNN to protect privacy. It then switches to uninterrupted execution of the later layers of DNN when the input reconstruction is no longer feasible. 
    \item Origami works on a pre-trained model and does not require any model changes or re-training to achieve its goals. 
    \item This work designs a strong conditional generative adversarial network (c-GAN)  based framework with unlimited inference requests as an input reconstruction adversary. Using the c-GAN we verify that the privacy of the user input is protected.
    \item This work evaluates the Origami framework on Intel SGX using two unmodified and pre-trained DNN models, namely VGG-16 and VGG-19,  and demonstrates that up to 15x speedup in inference latency can be achieved compared to running the full model inside the SGX secure enclave. 
\end{enumerate}

Rest of the paper is organized as follows. Section 2 provides brief background for the paper. Section 3 describes the motivation and the two key ideas that lead to the design of the Origami inference framework. Section 4 describes the conditional GAN models to verify input privacy. Section 5 describes the implementation details. Section 6 presents the experimental evaluation. Section 7 presents the related work and section 8 concludes the paper.


\section{Background}
\label{sec:background}

\begin{figure*}[h]
\centering
\begin{subfigure}{0.45\textwidth}
    \includegraphics[width=8cm, height=6cm]{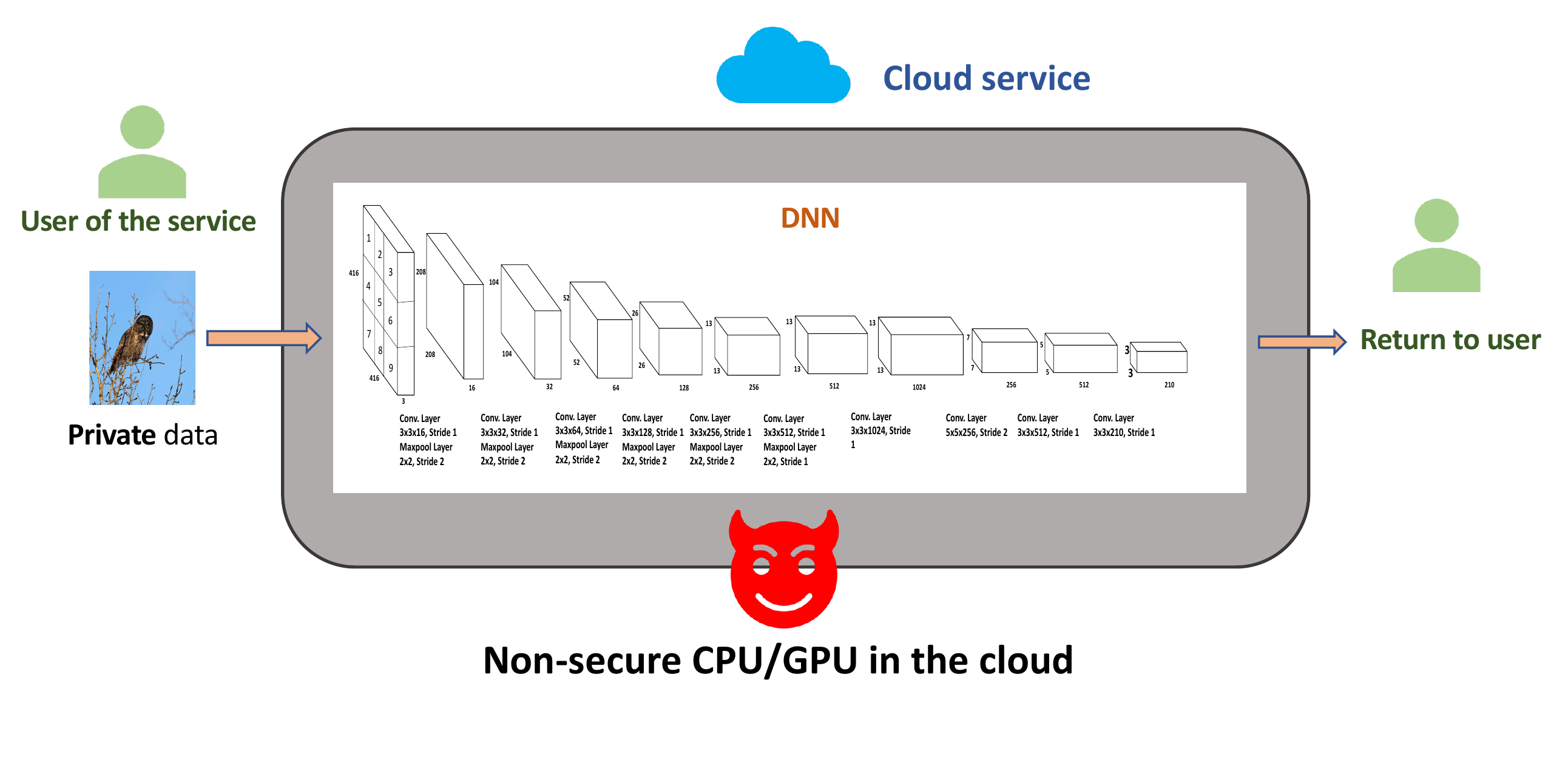}\\
    \caption{Unsecure system}
    \label{fig:systemModel_1}
\end{subfigure}
\begin{subfigure}{0.45\textwidth}
    \includegraphics[width=8cm, height=6cm]{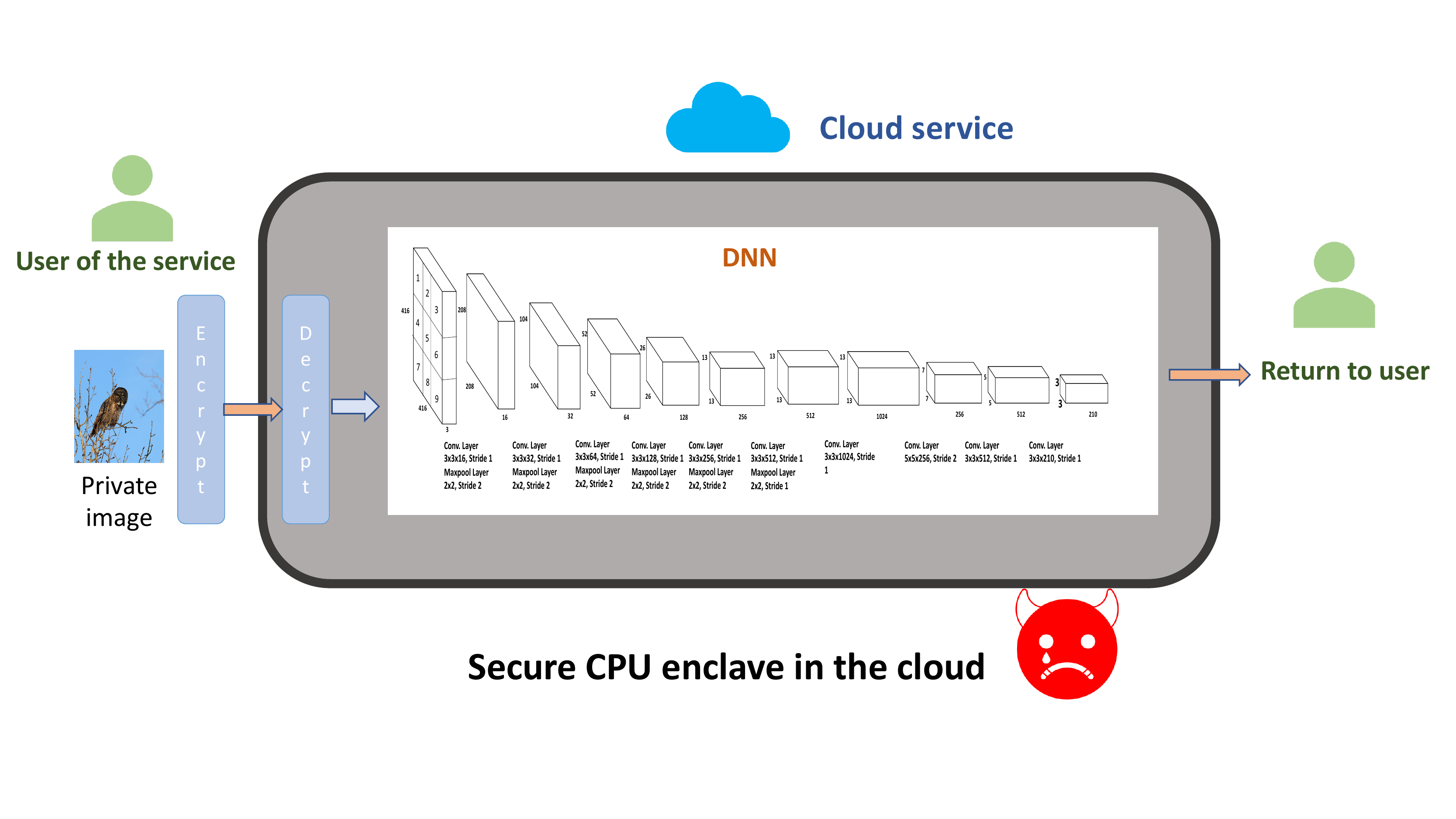}\\
    \caption{Secure system}
    \label{fig:systemModel_2}
\end{subfigure}
\end{figure*}

\subsection{TEEs and Intel SGX}

Trusted Execution Environments (TEEs) such as Intel SGX (Software Guard Extensions), ARM TrustZone \cite{inteldm, Costan2016IntelSE} enable execution of programs in secure hardware enclaves. They protect the confidentiality and integrity of the code and data that are executed inside the enclaves. In Intel’s  SGX a region of memory (128 MB by default) is reserved only for enclaves and memory accesses to this region are restricted by CPU. Only the instructions executing inside enclave can access this memory region. Non-enclave code can initialize the enclave via SGX-related CPU instructions like ECREATE, EENTER, EADD etc. Operating systems, root users and all other applications running on the same machine are prevented from accessing this memory region. Intel SGX provides support for remote attestation of an initialized enclave. This can be used by a remote party to verify the code and data inside the enclave immediately after its initialization. While there have been some demonstrated side channel attacks on SGX \cite{sgxPectre, foreshadown}, in this work we assume  that SGX execution can be secured using some of the recently published schemes \cite{stt,InvisiSpec}. 

\section{Origami: Motivation and Two Key Ideas}
\label{sec:framework}
\subsection{Overheads of secure execution}
To motivate the need for private inference, consider a cloud service which is used by a health care provider to classify medical images of patients. The health care provider (user of the service) sends private data (an image related to a patient) that may be processed using the system depicted in figure \ref{fig:systemModel_1}. 
Even if the user encrypts and sends the data, it needs to be decrypted before processing. The user image becomes visible to the service and it is the responsibility of the service to ensure confidentiality of the user data. The services in cloud are exposed to a wide attack surface like malicious cloud users, compromised hypervisors, physical snooping. 

TEEs like Intel SGX can be used to provide confidentiality since they support remote attestation and are designed to protect the confidentiality of the data running inside the secure enclaves.
Consider the service using a TEE based system shown in figure \ref{fig:systemModel_2}. In this system after the cloud service initializes the secure enclave, a user can get remote attestation that the initialized enclave loaded the proper code to process the user data. We assume that the model provided by the cloud service is verifiable by the user before using the service, or the user of the service may load their own trusted models for cloud execution.
Then the user encrypts the image, before sending it to the cloud service. The service processes the encrypted image completely inside a secure enclave. 
However, the inference latency of this service  can be an one or even two orders of magnitude longer than the first approach. 

\begin{figure}
\centering
  \includegraphics[width=8cm, height=6cm]{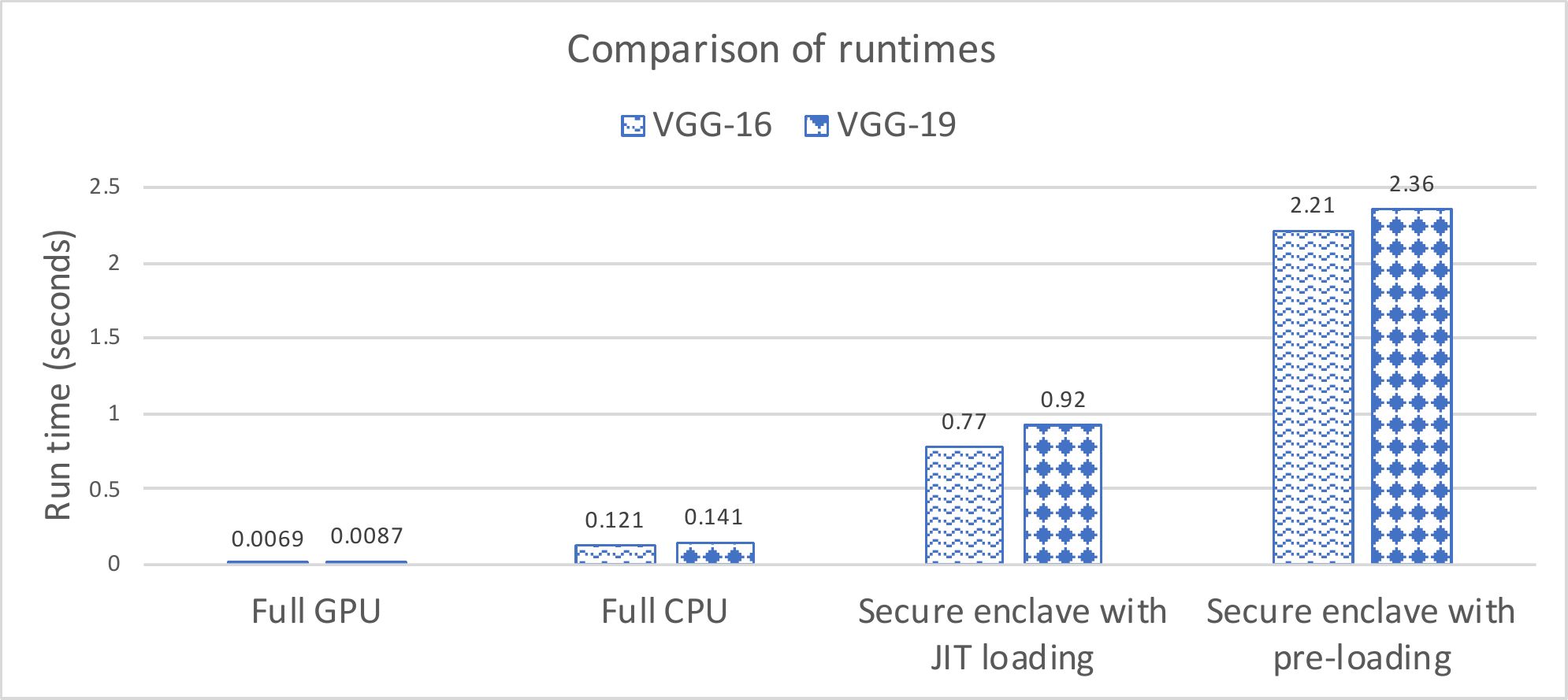}\\
  \caption{Comparison of runtimes}
  \label{fig:motivation}
\end{figure}

\begin{figure*}[h]
\begin{subfigure}{\textwidth}
\centering
  \includegraphics[height=8cm]{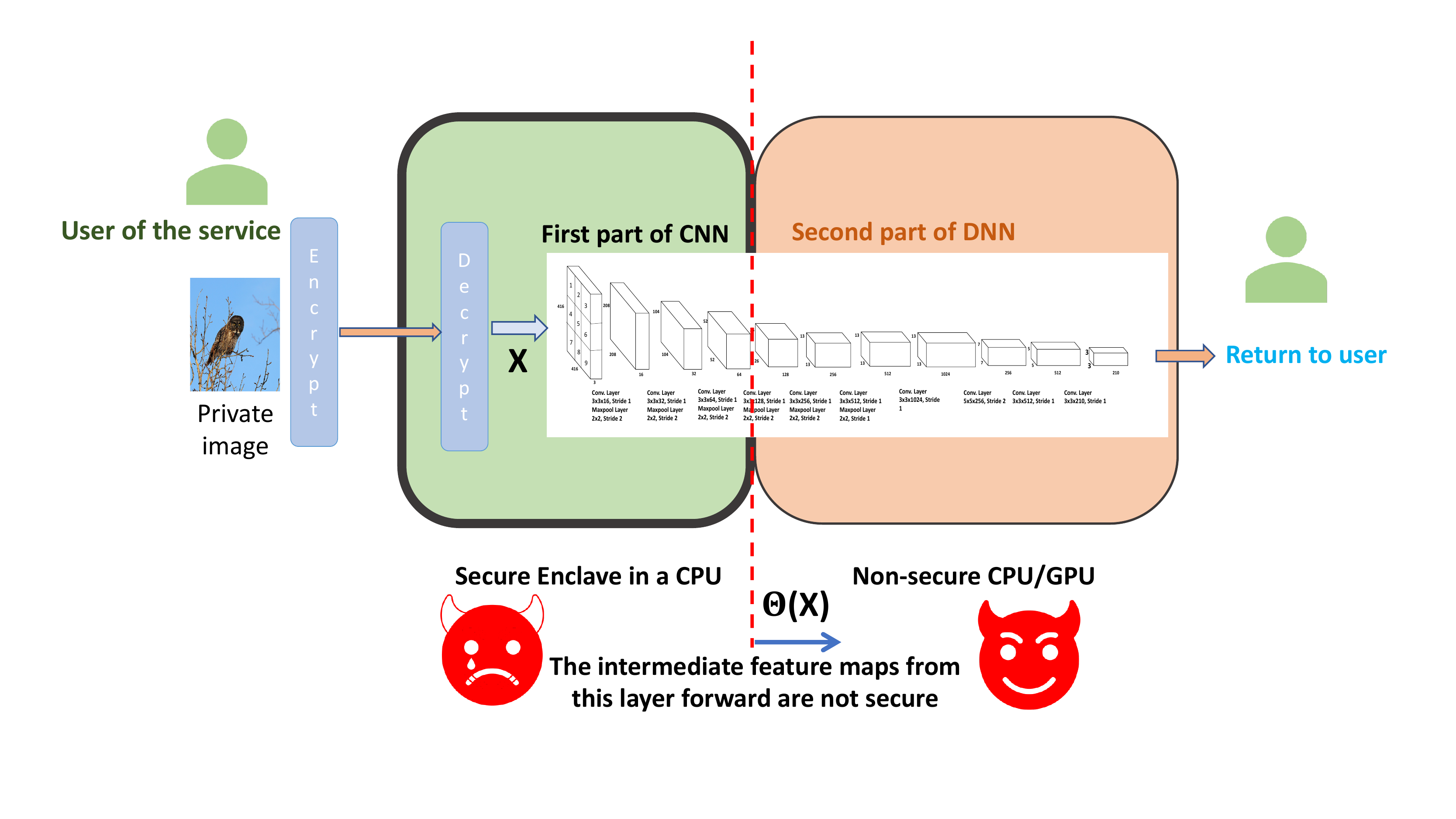}\\
  \caption{Secure system with partitioning}
  \label{fig:systemModel_3}
\end{subfigure}
\end{figure*}

In figure \ref{fig:motivation} we compare the inference runtimes of VGG-16 and VGG-19 models executing on an unsecure CPU with two secure enclave configurations (experimental setup details are shown later). In the first secure enclave configuration the model data is loaded Just-In-Time (JIT), only when needed, and in the second enclave configuration all the model data is pre-loaded. When compared to executing on a CPU with no privacy, the VGG-16 model runs 18.3x slowly on a secure enclave with pre-loading of data and 6.4x slowly on a secure enclave with JIT loading of data. Along the same lines, the VGG-19 model runs 16.7x slowly on a secure enclave with pre-loading of data and 6.5x slowly on a secure enclave with JIT loading of data.  The slowdowns are more drastic,  when compared to executing the model entirely on an untrusted GPU -- up to 321 times slower.  Eliminating or reducing this runtime slowdown on secure enclaves motivates Origami inference to consider model partitioning. 

\subsection{Key Idea1: Model partitioning and offloading}
Consider a service using a TEE based system with model partitioning as shown in figure \ref{fig:systemModel_3}. For example, with VGG-16, we propose that the 16 layers be split into two tiers $L$, 16 - $L$ layers. Prior to deploying the model in the cloud, the first tier with L layers is embedded within the SGX enclave container. The second tier of $16 - L$  layers is created as a separate container that can be executed in an open compute environment inside the CPU or can be offloaded to an accelerator. 

During operation, each inference request is encrypted by the user of the service and sent to the cloud. The encrypted input is then received by the first tier in the SGX enclave which securely decrypts the input. The input is then processed in the first $L$ layers. The output from the first tier is an intermediate feature map. The intermediate representation is then fed to the second tier which may be executed on a GPU/CPU. As we show in the next section, by appropriately selecting the  $L$ layers for first tier,  our model partitioning approach provides strong  guarantees that prevent input reconstruction from the intermediate representation generated by the first tier. Thus model partitioning can protect input privacy while also reducing the amount of computation performed within the SGX enclave. 



\begin{figure}
\centering
  \includegraphics[width=8cm, height=6cm]{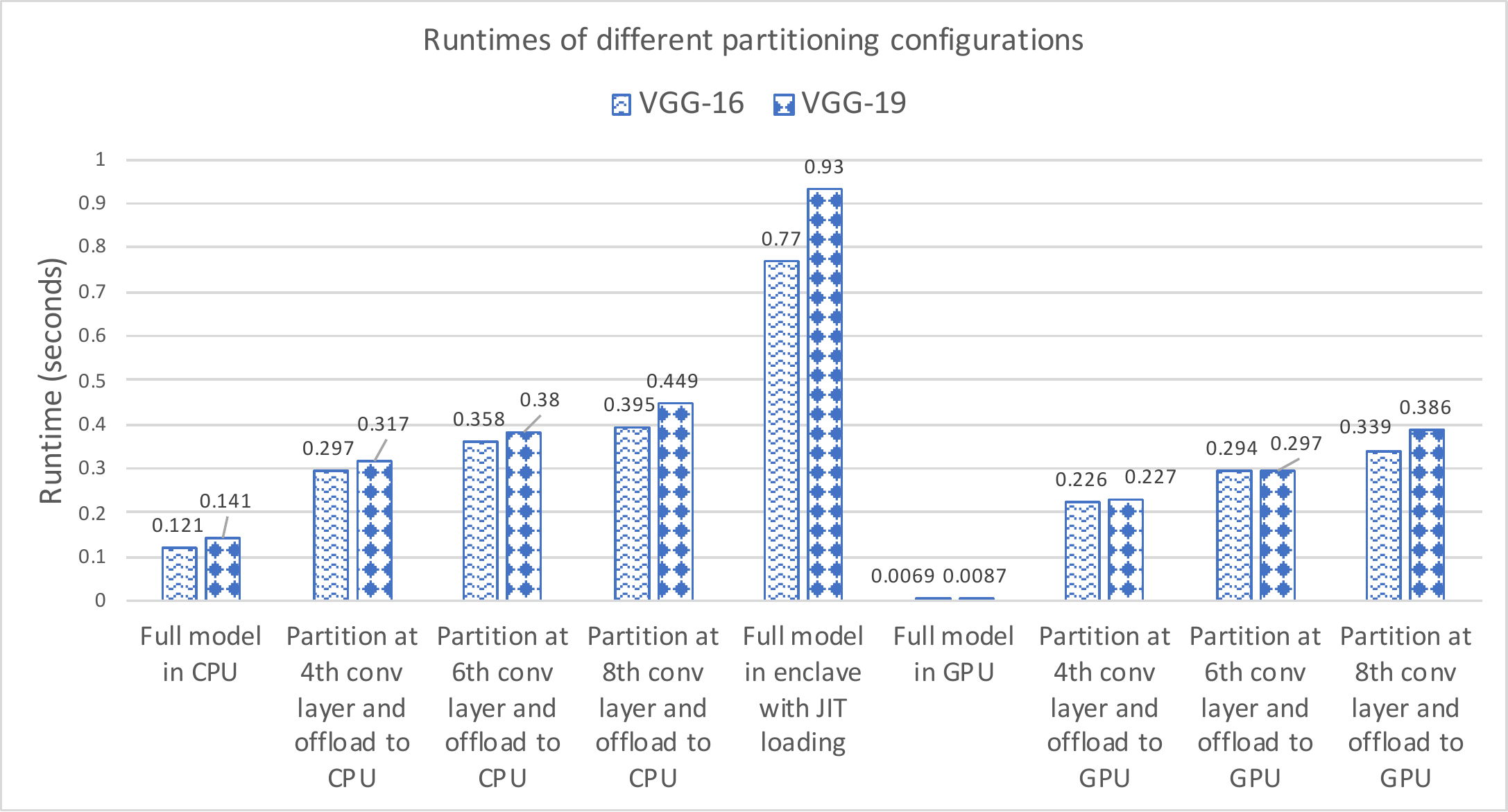}\\
  \caption{Runtime variation with different partitioning points}
  \label{fig:motivation_partitioning}
\end{figure}

We measure the inference runtime of the VGG-16 and VGG-19 models partitioned at different layers and plot these in figure \ref{fig:motivation_partitioning}. The first partition of the model is run inside SGX enclave and the second partition is run on a CPU outside of SGX. For VGG16, as partitioning point moves from 4th, 6th, to 8th convolutional layer, the inference slowdowns are 2.5x, 3x, and 3.3x respectively. For the VGG-19 model the slowdowns are 2.3x, 2.7x, and 3.2x respectively. More critically if the second partition is offloaded to a GPU, the slowdowns drop to about 50x, compared to up to 321x slowdown seen with GPU only execution. Although model partitioning considerably improves performance there is still a significant penalty that must be paid for private inference.   



\subsection{Key Idea2: Reducing SGX execution further with Slalom}
While model partitioning reduces the amount of computing done within SGX there is still a significant opportunity to lower the overheads. Slalom \cite{Tramr2018SlalomFV} proposed an approach to enable compute intensive operations to be offloaded to a GPU from SGX while still preserving privacy. Each layer of the DNN is split partially between GPU and SGX enclave. The compute intensive convolutions (matrix multiplications) within a DNN layer are offloaded to a GPU while allowing the non-linear operations (such as RELUs) to be performed within an enclave. However, offloading the convolution layers exposes the user input to an adversary. For instance, the first layer in a CNN model convolves the input image with a sliding window of feature matrix. To prevent the private data exposure Slalom uses a technique based on cryptography to blind the data before sending it to the GPU. This technique adds noise i.e., random elements, which is kept private within the SGX, to the input matrices within the enclave before releasing the data to the GPU. These random elements are referred to as the blinding factors. This noisy input data, with privately held blinding factors, creates the privacy guarantees by preventing an adversary from observing (or even reconstructing) the inputs. 


The GPU sends the results of its computation on the noisy data back to the SGX enclave. Because the GPU is performing only linear operations (matrix multiplications, in particular) one can decode GPU results by subtracting the precomputed noisy components.  For this purpose, the Slalom uses the privately stored unblinding factors to extract the computation result sans noise before applying non-linear functions within the enclave.   Slalom's reliance on decodability of computed data requires the approach to run only linear operations on GPU while requiring SGX to perform all non-linear operations. Non-linear operations on noisy data will essentially render the results undecodable. Thus Slalom \textit{must} pay the cost of blinding and unblinding every layer within a CNN model. 

We analyzed  the performance of Slalom compared to an execution without any privacy GPU. As we show in our result section (see Figure~\ref{fig:CPGPU}), the performance of Slalom is about 10x slower. Hence, even though Slalom offloads most of its computations to a GPU, it still pays non-trivial overheads. We analyzed the reasons for this slowdown. Our experiments showed that unblinding or blinding 6MB features roughly takes 4 milliseconds and there are roughly 47MB and 51MB intermediates features to process per each inference in VGG-16 and VGG-19. Hence, a significant fraction of the total execution time is hobbled by the encoding and decoding of data.  


\subsection{Origami: Combining model splitting with blinding}
Using cryptographic blinding can eliminate the cost of SGX overheads but leads to an increase in blinding and deblinding overheads. These blinding overheads can be eliminated if one can avoid blinding the data once the input reconstruction capability is no longer a concern. By avoiding blinding at the earliest possible DNN layer it is possible to execute even the non-linear operations within a GPU thereby allowing an uninterrupted execution of multiple CNN layers within a GPU. Origami framework combines both the techniques. It first outsources only the linear operations of the layers within the first partition while  executing the non-linear operations inside the enclave.  It then completely outsources the second partition of a DNN for execution on a CPU or GPU.



\section{Adversary and Threat model}
\label{sec:reconstruction}
Offloading the second partition of the model to an unsecure CPU or GPU makes all the intermediate data of these layers available to adversaries. An adversary in the Origami framework is an agent that tries to use the observed intermediate data of the model to reconstruct the input image sent by the user to the service. Origami makes the assumption that model weights inside the enclave are private and protected but the intermediate results from the first tier are openly visible to an adversary. More specifically, let the function computed by the hidden layers inside the enclave be $\Theta$, as shown in figure \ref{fig:systemModel_3}. For some input $X$, an adversary can observe $\Theta(X)$ and then must find the optimal $X'$ that minimizes the loss between $\Theta(X)$ and $\Theta(X')$ \cite{inverting_whitebox}.

\begin{figure}
\centering
  \includegraphics[width=8cm, height=6cm]{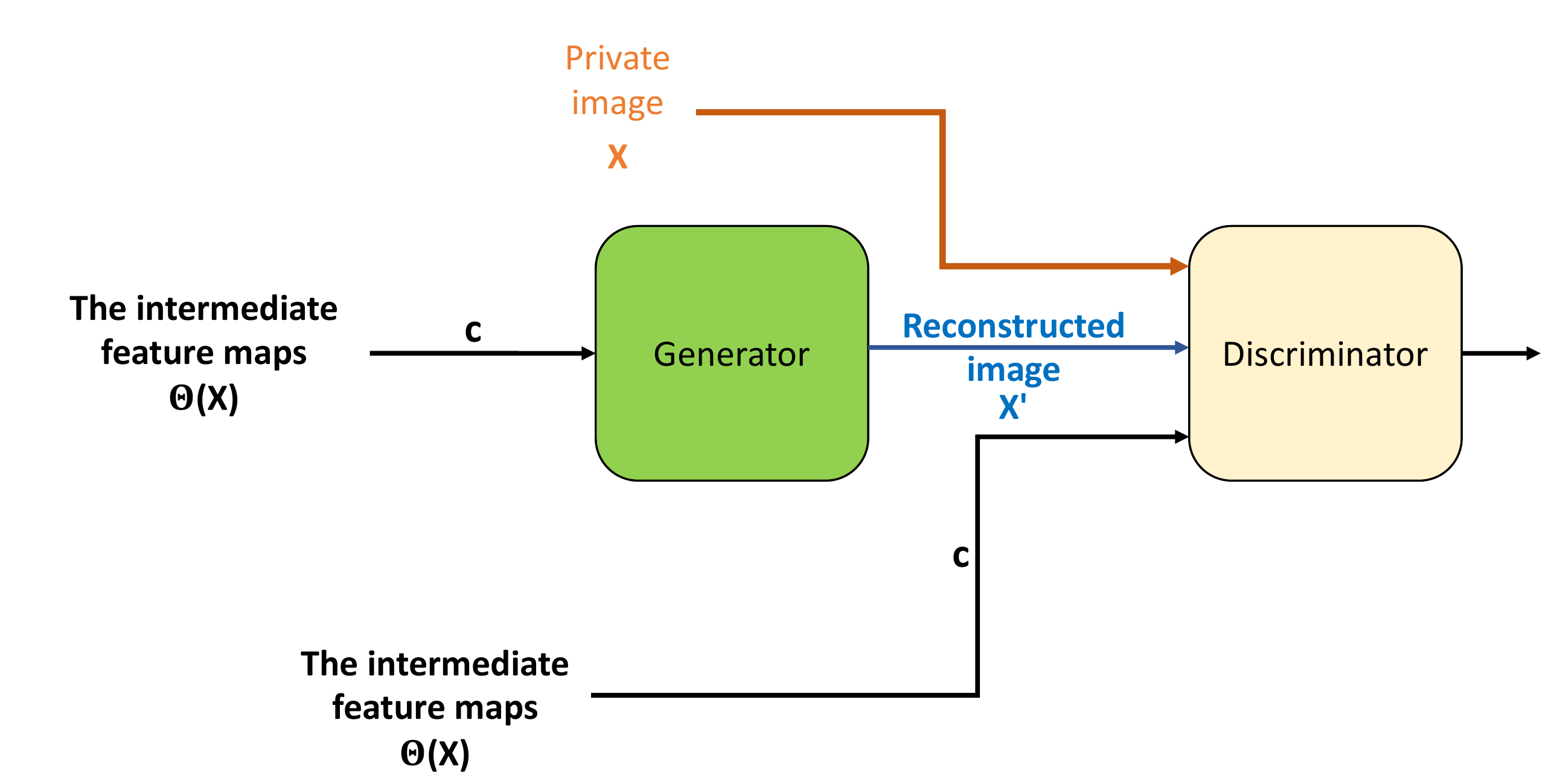}\\
  \caption{c-GAN adversary model}
  \label{fig:gan1}
\end{figure}

\subsection{C-GAN Adversary}  
Origami inferences uses a strong adversary model shown in figure \ref{fig:gan1}, which is a conditional Generative Adversarial Network (c-GAN) \cite{DBLP:journals/corr/MirzaO14} to learn $\Theta$. GAN \cite{Goodfellow2014GenerativeAN} is a network that consists of two DNN models that evolve together, a generator model $G$ and a discriminator model $D$. The generator model $G$ learns the training data distribution by striving to generate "fake" samples that are difficult for the discriminator $D$ to differentiate from the real samples; at the same time the discriminator model D learns to distinguish if a given sample is coming from the training data or synthesized by the generator $G$. The training process for both models resembles a two-player min-max game to optimize for the following objective function, where $x$ is from the training dataset distribution and $z$ is noise sampled from a multi-variate Gaussian distribution:
\begin{equation}
\min\limits_{G}\max\limits_{D}\{E_{x\sim pdata}[\log D(x)] + E_{z\sim pz}[\log (1-D(G(z))]\}
\end{equation}
Conditional GAN \cite{DBLP:journals/corr/MirzaO14} builds on top of GAN and uses additional data $c$ to condition both the generator and discriminator models, yielding $G(z, c)$ and $D(x, c)$. This extension enables $G$ to generate samples conditioned on some variable $c$.

Recall that in Origami the first tier uses blinding factors to preserve privacy,  while the second tier of computing is done entirely in the open. Thus in the context of Origami, the adversary trains a c-GAN on the intermediate data generated from the first tier. The generator network G is trained to produce the private image inside the enclave as its output. It takes $\Theta(X)$ as input and trains to generate a $X'$ as real as possible. A discriminator network is trained to classify between a real image $X$ and a fake image $X'$ produced by the generator, given $\Theta(X)$. After the c-GAN is trained, the adversary can use the generator to take in the intermediate data between the partitions and generate the private input image.

\subsection{Training the c-GAN adversary}  

The ability of an adversary to train the c-GAN model depends on two things
\begin{enumerate}
    \item The information about the input image that is contained in the intermediate data.
    \item The quantity of the training data.
\end{enumerate}

Collecting a lot of training data requires the adversary to send a huge number of queries to the service and observe all the corresponding intermediate data to collect enough [$\Theta(X)$, $X$] pairs. In this work we make the assumption that the adversary has significant resource advantage to pay for the cloud service and the cloud service will not limit the number of queries.  

Given these strong adversary assumptions, the intermediate data should not provide  information to enable reconstruction of the input images. Then even collecting a large amount of training data will not help the c-GAN model reach high accuracy. Consider a limiting example of trying to reconstruct an input image using just the probabilities of the last layer of the CNN model. In this example all the layers are running inside the enclave and only the final softmax probabilites are available outside. It has been empirically shown that using the softmax probabilities cannot give a good reconstruction of the input image \cite{ImageInvertion}. The softmax probabilities can contain information like the color of the image but lack the information on reconstructing the exact objects in the image. Using earlier layers as partitioning points increases the probability of an adversary successfully reconstructing the user input whereas using deeper layers as partitioning points makes it infeasible for an adversary to successfully reconstruct the user input.

\subsection{Model partitioning algorithm} 
Origami framework finds the partitioning layer where splitting the model makes it empirically  infeasible for the adversary to perform a good reconstruction of the image. The procedure followed is described in Algorithm \ref{algo:partition}. Beginning from the first layer of the model the procedure empirically verifies, by training a c-GAN, if input images can be reconstructed from the intermediate feature maps that are outputs of the current layer. The metric used to measure the reconstruct-ability is the popular Structural Similarity (SSIM) \cite{ssim} between the real images, $X$, and the reconstructed images, $X^{'}$. SSIM metric measures the structural similarity between two images and is based on the perception of human visual system. A value of 0 indicates no structural similarity.


One surprising observation we found is that  if our algorithm finds that the feature maps from a layer $p$ cannot reconstruct the inputs, the c-GAN can reconstruct input from a deeper layer. Thus it becomes necessary to verify c-GAN capabilities for deeper layers. So, the procedure verifies reconstruct-ability at layers $p+1$ and $p+2$ etc., For example, in a VGG-16 model we observed that it is infeasible for the c-GAN to reconstruct the input images by using feature maps from the first max pool layer. However, using the feature maps from the convolutional layers that follow the max pool layer it is feasible to reconstruct the input images. In our understanding this happens because the feature maps from the max pool layer lack the spatial relationships. However, feature maps in the convolutional layers that follow the first max pool layer seem to recover enough spatial relationships to get a good reconstruction of the input images. We provide more details on the reconstruct-ability in the evaluation section.

\begin{algorithm}[h]
\caption{Model partitioning algorithm}
\label{algo:partition}
\begin{algorithmic}
\State \textbf{Input:} Model $M$ with $L$ layers, c-GAN architecture, image dataset
\State \textbf{Output:} Partitioning layer $p$
\For{each layer $l$ starting from first layer in $L$}
       \State Collect the intermediate feature maps $maps_{i}$ that are the
       \State outputs of layer $l$ for every image $i$ in the dataset
       \State Train a c-GAN using all the ($i$, $maps_{i}$) pairs
       \State Check for reconstruct-ability using SSIM metric
\EndFor
\State Let $p$ be the layer such that it's feature maps $maps_{p}$ 
\State cannot be used by the c-GAN to reconstruct the corresponding input images
\State Verify reconstruct-ability using maps of next two layers
\State $p+1$, $p+2$
\If {reconstruct-ability is infeasible for $p+1$, $p+2$}
      \State select $p$ as the partitioning point for the model $M$
\Else
      \State Go back to the \textit{for} loop and start from layer $p+1$
\EndIf
\end{algorithmic}
\end{algorithm}

\section{Implementation}
\label{sec:evaluation}
\subsection{c-GAN architecture and implementation}

\begin{figure}
\centering
  \includegraphics[width=8cm, height=6cm]{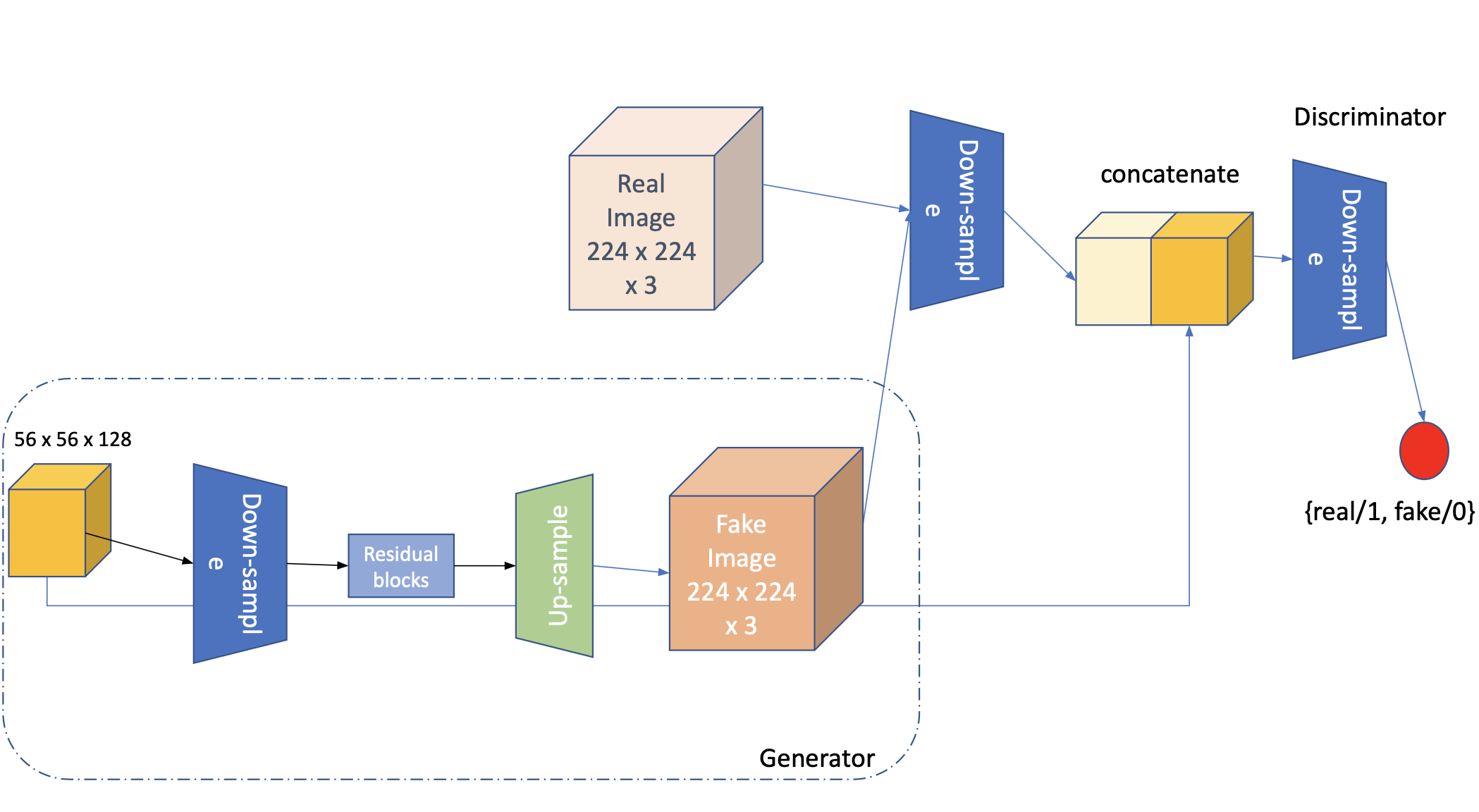}\\
  \caption{c-GAN architecture}
  \label{fig:gan-arch}
\end{figure}

We design the Generator $G$ of the GAN model as an encoder-decoder network with a series of residual blocks \cite{resnet}. As shown in figure \ref{fig:gan-arch}, the intermediate feature map output taken from the partition layer $p$ is first fed into the encoder, which is composed of 3 convolutional layers to down-sample until it has a spatial dimension of \(14  \times 14\). Then this encoded feature map is fed into 4 residual blocks, before going through a series of 4 up-sampling blocks acting as a decoder to generate a fake image of dimension \(224 \times 224 \times 3\). The residual block consists of \(3 \times 3\) stride 1 convolution, batch normalization and ReLU. The up-sampling block consists of nearest-neighbor upsampling followed by a \(3 \times 3\) stride 1 convolution. Batch normalization and ReLU are applied after each convolutional layer.

On the other hand, the Discriminator network $D$ first takes in an input image of dimension \(224 \times 224 \times 3\) and let it go through 2 convolutional layers to down-sample until it has the same spatial dimension as its intermediate feature map $c$ produced from the partition layer $p$ in VGG-16. Then it is concatenated together with the intermediate feature map $c$ as condition to be fed into a series of 5 consecutive convolutional layers, and then a fully connected layer with Sigmoid activation, to predict whether the input image is "true" i.e., it is from the training dataset, or "fake" generated by the Generator network $G$. Here the convolutional layers are \(4 \times 4\) with stride 2 and except the last one layer, they are all followed by batch normalization and LeakyReLU with 0.2 negative slope.

The architectures of the Generator and Discriminator in the c-GAN are tuned, as needed, to work with the different shapes of the intermediate feature maps from different partitioning layers. The c-GAN models, the code to perform model partitioning and collect the intermediate feature maps are written in Python language using Keras library.

We train the c-GAN adversary models using images from the Imagenet ILSVRC 2012 validation dataset \cite{deng2009imagenet}. This dataset comprises 50,000 images belonging to 1000 classes. To check reconstruction at each partitioning layer, the c-GAN is trained for 150 epochs on a single NVIDIA GTX 1080 Ti GPU with a learning rate of 0.0002. The training time varies depending on the size of the feature maps of the partitioning layer. Training for 150 epochs for the 2nd layer took nearly seven days, and for the 7th layer it took nearly one and half days.

\begin{figure*}[h]
\centering
\begin{subfigure}{0.45\textwidth}
  \centering
  \includegraphics[width=8cm, height=2cm]{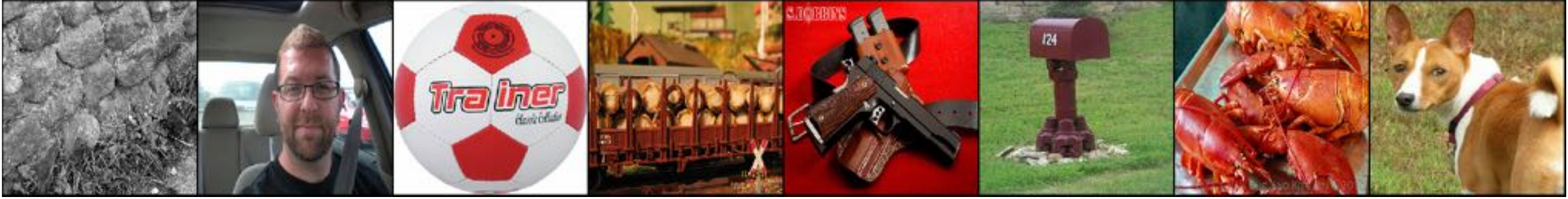}\\
  \caption{Real images}
  \label{fig:vgg16_layer1_real}
\end{subfigure}
\begin{subfigure}{0.45\textwidth}
\centering
  \includegraphics[width=8cm, height=2cm]{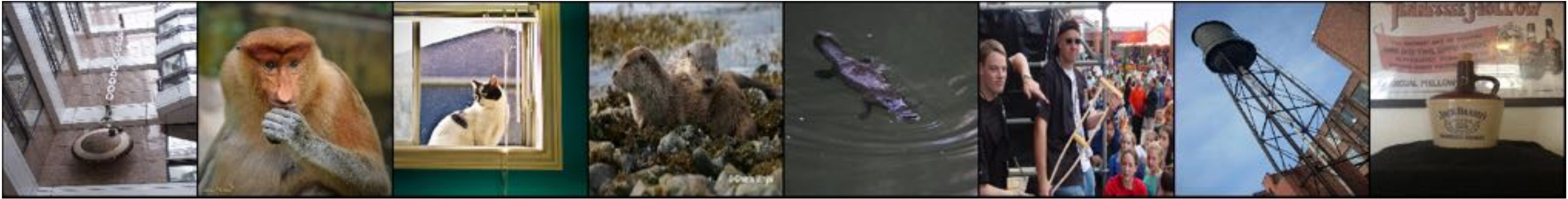}\\
  \caption{Real images}
  \label{fig:vgg16_layer1_generated}
\end{subfigure}
\begin{subfigure}{0.45\textwidth}
\centering
  \includegraphics[width=8cm, height=2cm]{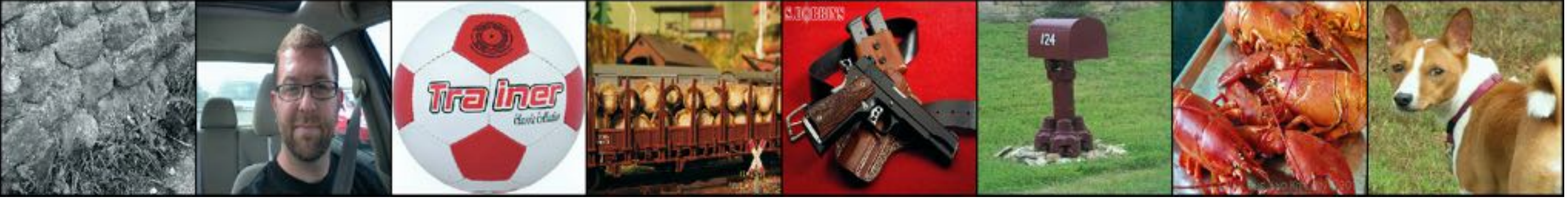}\\
  \caption{Reconstructed images from layer 1 (conv) feature maps}
  \label{fig:vgg16_layer1_generated}
\end{subfigure}
\begin{subfigure}{0.45\textwidth}
\centering
  \includegraphics[width=8cm, height=2cm]{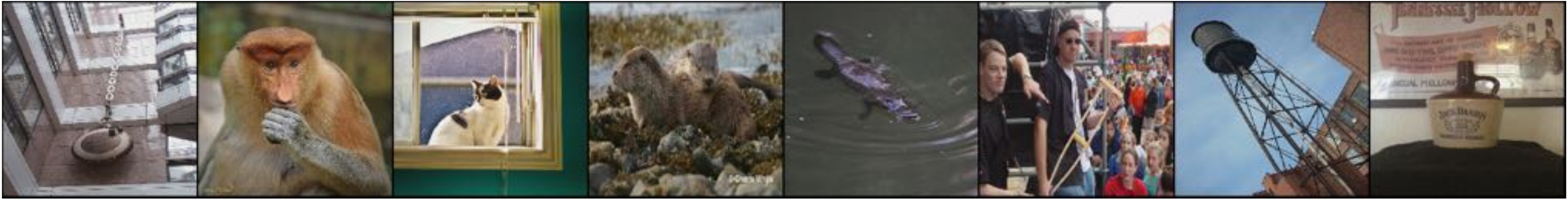}\\
  \caption{Reconstructed images from layer 2 (conv) feature maps}
  \label{fig:vgg16_layer1_generated}
\end{subfigure}
\begin{subfigure}{0.45\textwidth}
  \centering
  \includegraphics[width=8cm, height=2cm]{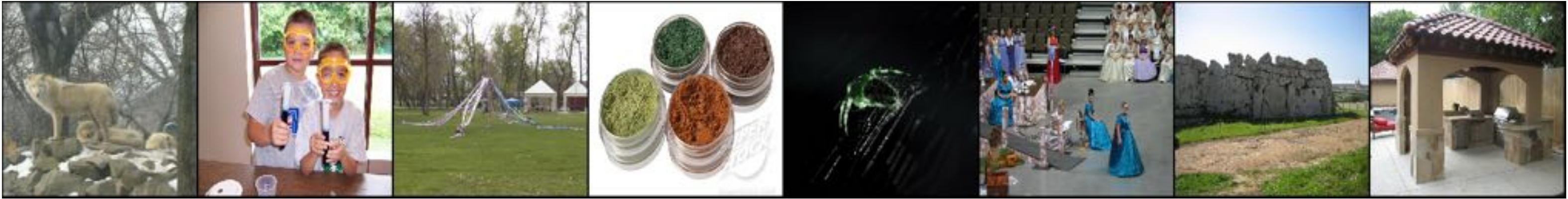}\\
  \caption{Real images}
  \label{fig:vgg16_layer1_real}
\end{subfigure}
\begin{subfigure}{0.45\textwidth}
\centering
  \includegraphics[width=8cm, height=2cm]{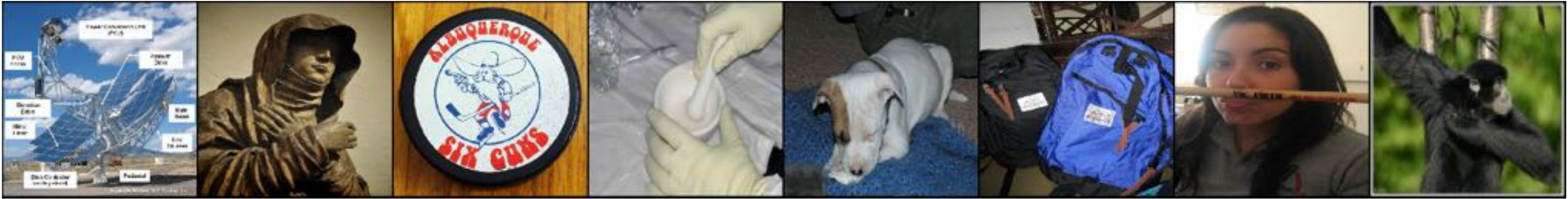}\\
  \caption{Real images}
  \label{fig:vgg16_layer1_generated}
\end{subfigure}
\begin{subfigure}{0.45\textwidth}
\centering
  \includegraphics[width=8cm, height=2cm]{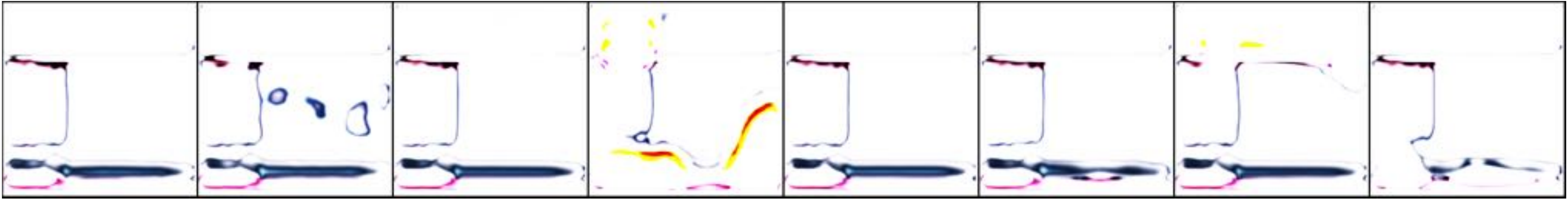}\\
  \caption{Reconstructed images from layer 3 (max pool) feature maps}
  \label{fig:vgg16_layer1_generated}
\end{subfigure}
\begin{subfigure}{0.45\textwidth}
\centering
  \includegraphics[width=8cm, height=2cm]{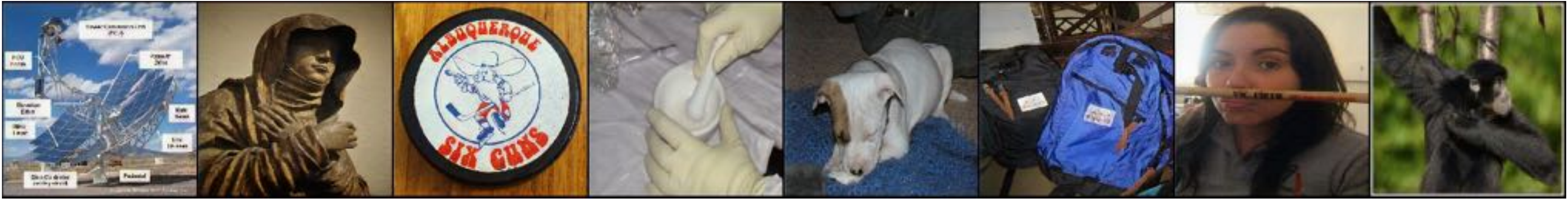}\\
  \caption{Reconstructed images from layer 4 (conv) feature maps}
  \label{fig:vgg16_layer1_generated}
\end{subfigure}
\begin{subfigure}{0.45\textwidth}
  \centering
  \includegraphics[width=8cm, height=2cm]{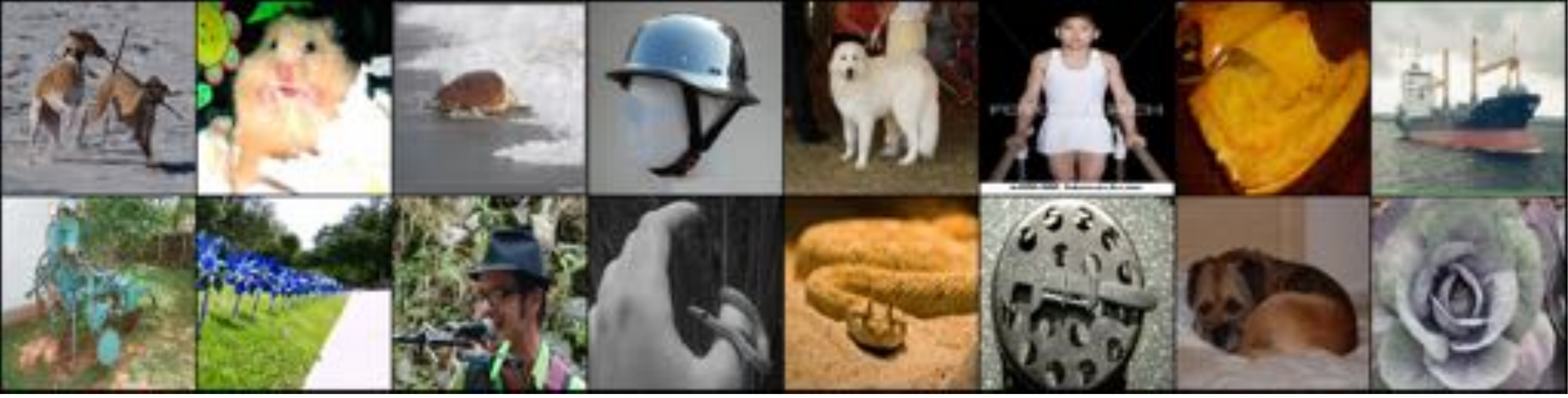}\\
  \caption{Real images}
  \label{fig:vgg16_layer1_real}
\end{subfigure}
\begin{subfigure}{0.45\textwidth}
\centering
  \includegraphics[width=8cm, height=2cm]{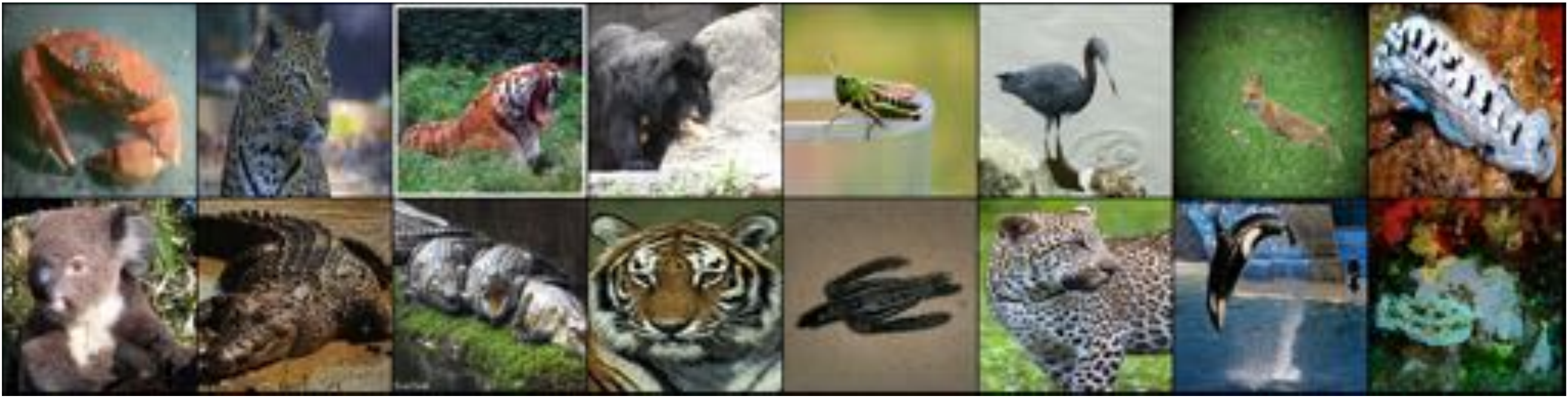}\\
  \caption{Real images}
  \label{fig:vgg16_layer1_generated}
\end{subfigure}
\begin{subfigure}{0.45\textwidth}
\centering
  \includegraphics[width=8cm, height=2cm]{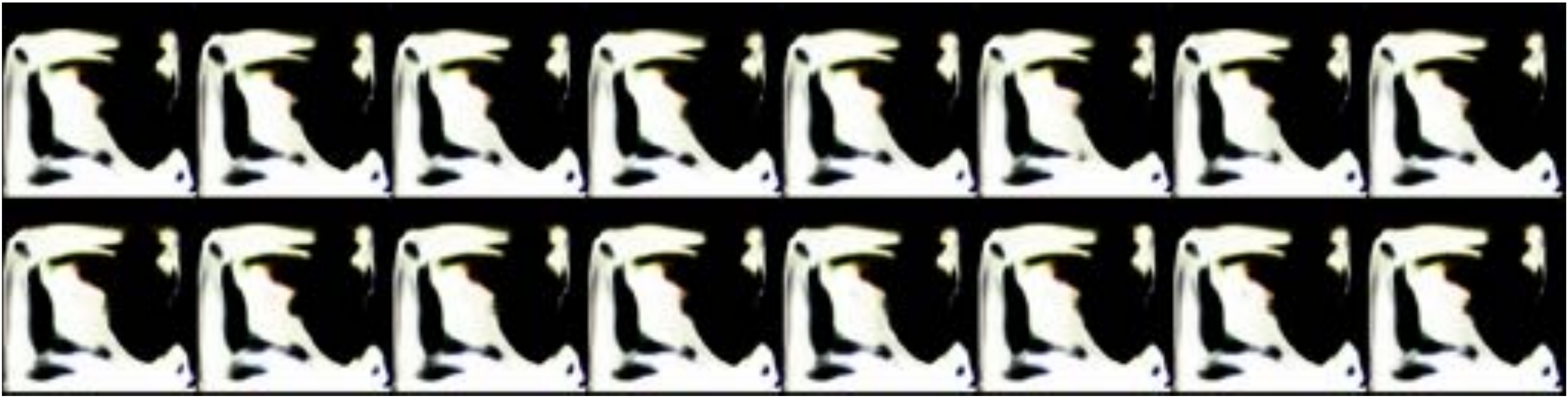}\\
  \caption{Reconstructed images from layer 6 (max pool) feature maps}
  \label{fig:vgg16_layer1_generated}
\end{subfigure}
\begin{subfigure}{0.45\textwidth}
\centering
  \includegraphics[width=8cm, height=2cm]{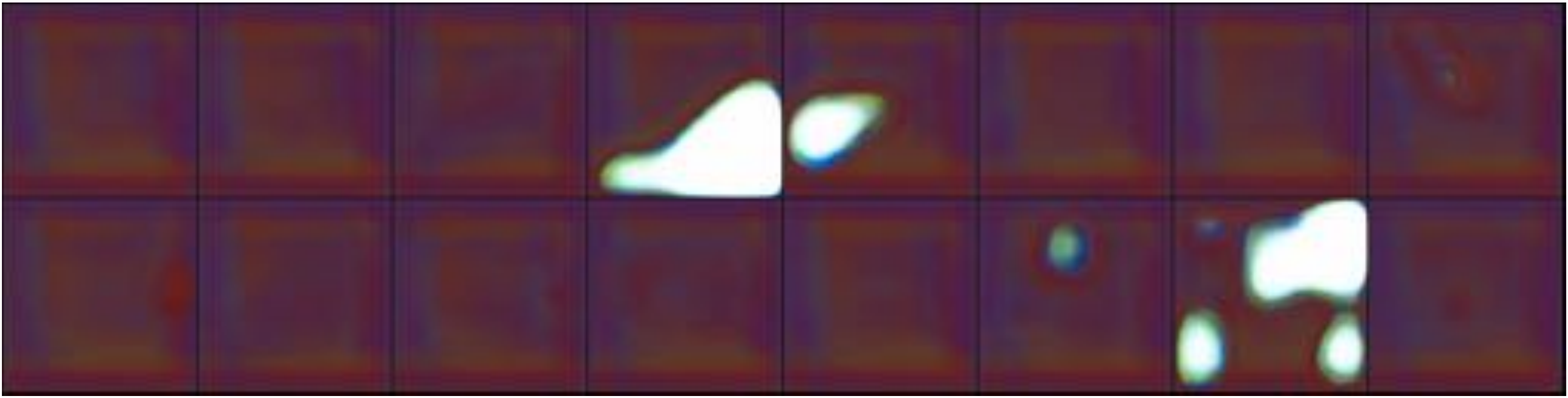}\\
  \caption{Reconstructed images from layer 7 (conv) feature maps}
  \label{fig:vgg16_layer1_generated}
\end{subfigure}
\caption{Real images and reconstructed images using feature maps of different layers in VGG-16}
\label{fig:vgg16_reconstruct_gan}
\end{figure*}

\subsection{Implementation of models in SGX}
The code to execute the VGG-16 and VGG-19 models inside SGX enclaves is written in Python and C++ using the SGXDNN library from Slalom. The code to execute the models on CPU or GPU is written using Keras library in Python.

All the models and SGX code is  open source and can be downloaded from GitHub at 
\url{https://github.com/cjbaq-origami/origami_inference}. 



\section{Evaluation}
\label{sec:evaluation}

Here is the outline of our evaluation presentation. As a first step, we experimentally evaluate the privacy guarantees of Origami inference, using c-GANs to reconstruct original images from intermediate features from the two-tier implementation of VGG. Second, we demonstrate the performance of Origami inference, against baseline strategies using Slalom's SGXDNN library \cite{Tramr2018SlalomFV}. We  compare our framework's performance with that on untrusted CPUs and GPUs to show how much overhead we still pay to protect privacy. 

\subsection{Hardware configuration}
We performed all our evaluations on a server consisting of a Intel Xeon E-2174G CPU equipped with SGX capability and a NVIDIA GeForce GTX 1080 Ti GPU attached as an accelerator. The CPU has 8 threads and 64 GB memory. It's operating system is Ubuntu 18.04. The GPU has 11 GB of GDDR memory.

\subsection{Partitioning and input privacy}
\begin{figure}
\centering
  \includegraphics[width=8cm, height=6cm]{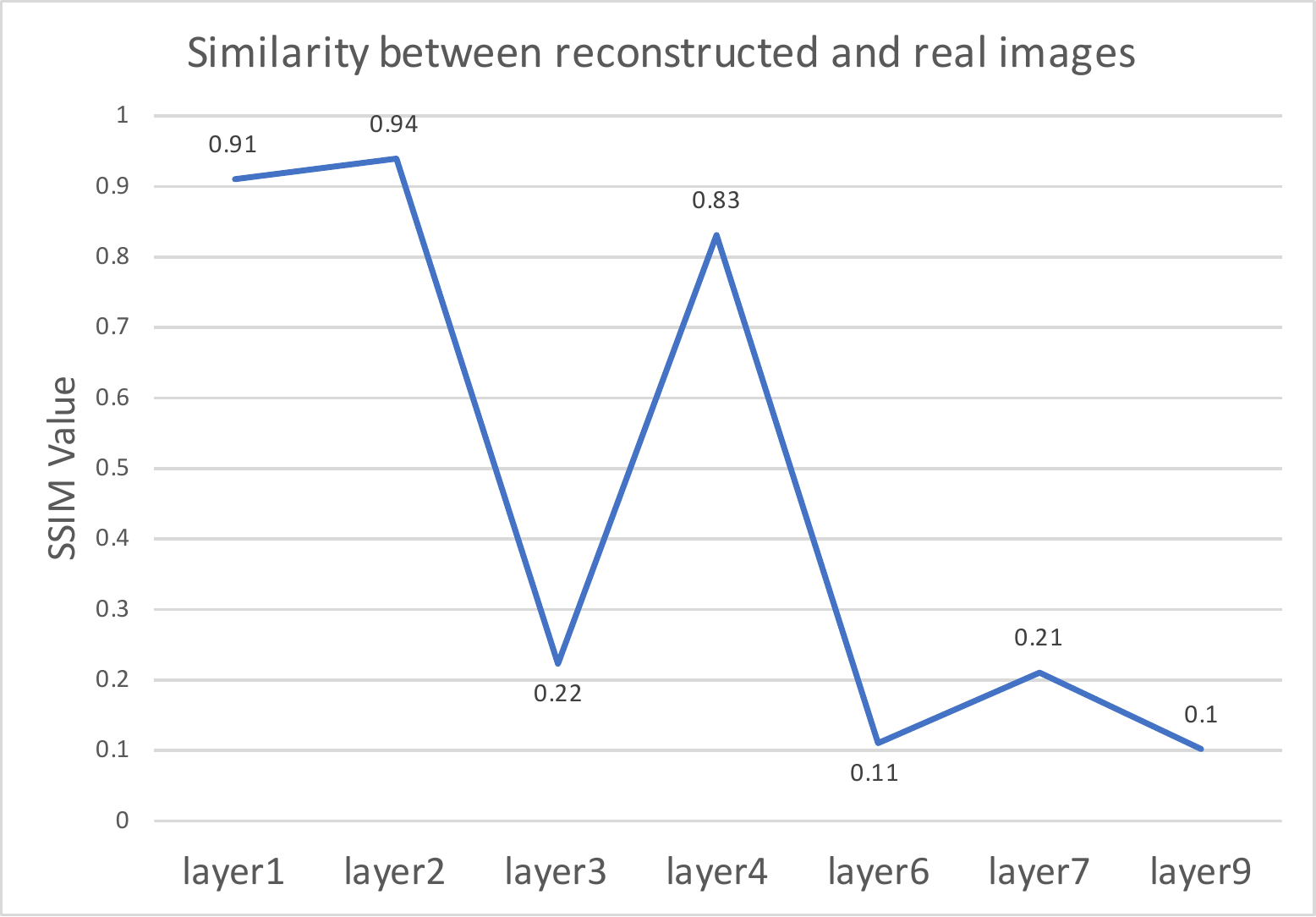}\\
  \caption{Similarity between real and reconstructed images at different partition layers}
  \label{fig:ssim_values}
\end{figure}

We train the c-GAN adversary models and evaluated the Origami framework using images from the Imagenet ILSVRC 2012 validation dataset \cite{deng2009imagenet}. Image reconstructions by the c-GAN using feature maps at different layers of the VGG-16 model, while running the model partitioning algorithm \ref{algo:partition}, are shown in figure \ref{fig:vgg16_reconstruct_gan}. Figures \ref{fig:vgg16_reconstruct_gan}(a) and (b) show a sample of real images sent for private inference. In (c) and (d) we show the reconstructed images generated from the intermediate outputs from a trained c-GAN. We show the reconstruction capability of c-GAN after the first and second convolution layers of VGG-16. In (e) and (f) we show another sample real images sent for private inference. In (g) and (h) we demonstrate the surprising result that the c-GAN is unable to reconstruct the input after layer 3, but it was able to reconstruct input image from layer 4. As discussed earlier, feature maps in the convolutional layers that follow the first max pool layer seem to recover enough spatial relationships to get a good reconstruction of the input images. In (i) and (j) we show real images and the reconstruction output from c-GAN using the intermediate outputs from layer 6 and 7. Clearly after layer 6, in VGG-16 the c-GAN is unable to reconstruct the input, even when the c-GAN is trained with all input images.

We show the average SSIM metric values between real and reconstructed images at different layers of the VGG-16 model in figure \ref{fig:ssim_values}. The variation of these values clearly indicates the visual similarity (or its lack thereof) between real and reconstructed images already shown in figure \ref{fig:vgg16_reconstruct_gan}. The visual similarity is high for the first 2 layers, drops significantly for the third layer, is again high for the fourth layer, then decreases and stays below 0.2 for all layers past layer 7.

\subsection{Performance Evaluation}
We use three metrics to measure performances of Origami:
(1) Inference runtime: Total time taken to perform a single inference. (2) SGX enclave memory requirement. SGX enclave memory is a precious resource and Intel SGX tool chain requires enclave writers to specify enclave memory usage statically. During application runtime, allocation of memory bigger than specified will trigger an exception. (3) Power event recovery time.  Intel's SGX capable processors will destroy the memory encryption keys inside them when a power event, such as hibernation, happens\cite{inteldm}. Thus, SGX applications need to recreate enclave after power events. Power Event recovery time is collected to measure Origami inference's ability to recover from unexpected power events. 

\noindent \textbf{Baseline Strategies:}
We compare Origami with several baselines. 
The baseline strategy  (referred to as Baseline2 in the figures) performs lazy loading of model parameters into SGX when loading fully connected layers that require more than 8MB memory. Parameters for such layers are loaded into enclave on demand. Although this strategy induces a small inference performance penalty by distributing parts of model loading cost into inference penalty, it can reduce the memory usage inside SGX enclave which avoids destructive page-swapping. Baseline 2 is also the baseline the original Slalom paper used. Note that we also evaluated another baseline (baseline 1) where all the model parameters are fully loaded before starting the inference but that baseline was much worse due to page swapping costs and hence was discarded. 

The Slalom/Privacy\cite{Tramr2018SlalomFV} model implements Slalom approach where \textit{all} the layers are interspersed between SGX and CPU/GPU. All linear computations are offloaded to GPU after applying blinding, and all blinding, unblinding and non-linear computations are performed within SGX.  Unblinding factors are pre-computed and are not part of the inference time. Blinding factors are generated on demand using the same Pseudo Random Number Generator seed while unblinding factors are encrypted and stored outside SGX enclave. When removing noise from intermediate features, Slalom/Privacy will only fetch parts of unblinding factors needed for a given layer into SGX enclave. This mode of operation is identical to the approach suggested in the original Slalom paper. 

We also evaluate three model splitting only strategies, where the first $L$ layers are executed within SGX and the remaining layers are sent to CPU/GPU. We refer to them as Split/\textit{x}, where \textit{x} is a positive integer signaling the \textit{x}th layer, after which all layers are offloaded to untrusted hardware. 

\subsubsection{SGX enclave Memory Usage}
Table \ref{table:1} shows the required SGX enclave memory, which is a major limiting factor for using SGX.  Baseline2 uses about 86MB memory, even though the VGG-16 model size more than 500 MB, due to the lazy loading process. But it may pay on-demand data loading penalty during inference. Origami (and Slalom) have about 2x lower memory overhead than this baseline. Slalom/Privacy requires 39MB SGX enclave memory, 12MB of which are used to temporarily store blinding/unblinding factors. Origami requires the same amount of SGX enclave memory as Slalom/Privacy does. Both of them have to accommodate enough blinding factors that can blind the largest intermediate feature map. The largest intermediate feature map for both Slalom/Privacy and Origami is about 12MB. Also compared to simple model splitting process Origami does pay a slight increase in memory penalty due to the use of blinding factors. 

There are two main benefits of reduced memory requirement. Our framework can retain its performance even with significantly low SGX enclave memory. Thus future models that are much more complex and may need much larger model sizes may be better accommodated in our framework.  Second, reduced memory requirement also allows more enclave applications to run simultaneously. Currently, the maximum available SGX memory on an Intel chip is 128MB. For Origami, there is still about 90MB free physical memory that can be used for other SGX enclave applications or to even co-run multiple private inference models.

\subsubsection{Inference Runtime}
\begin{figure}
\centering
  \includegraphics[width=8cm, height=6cm]{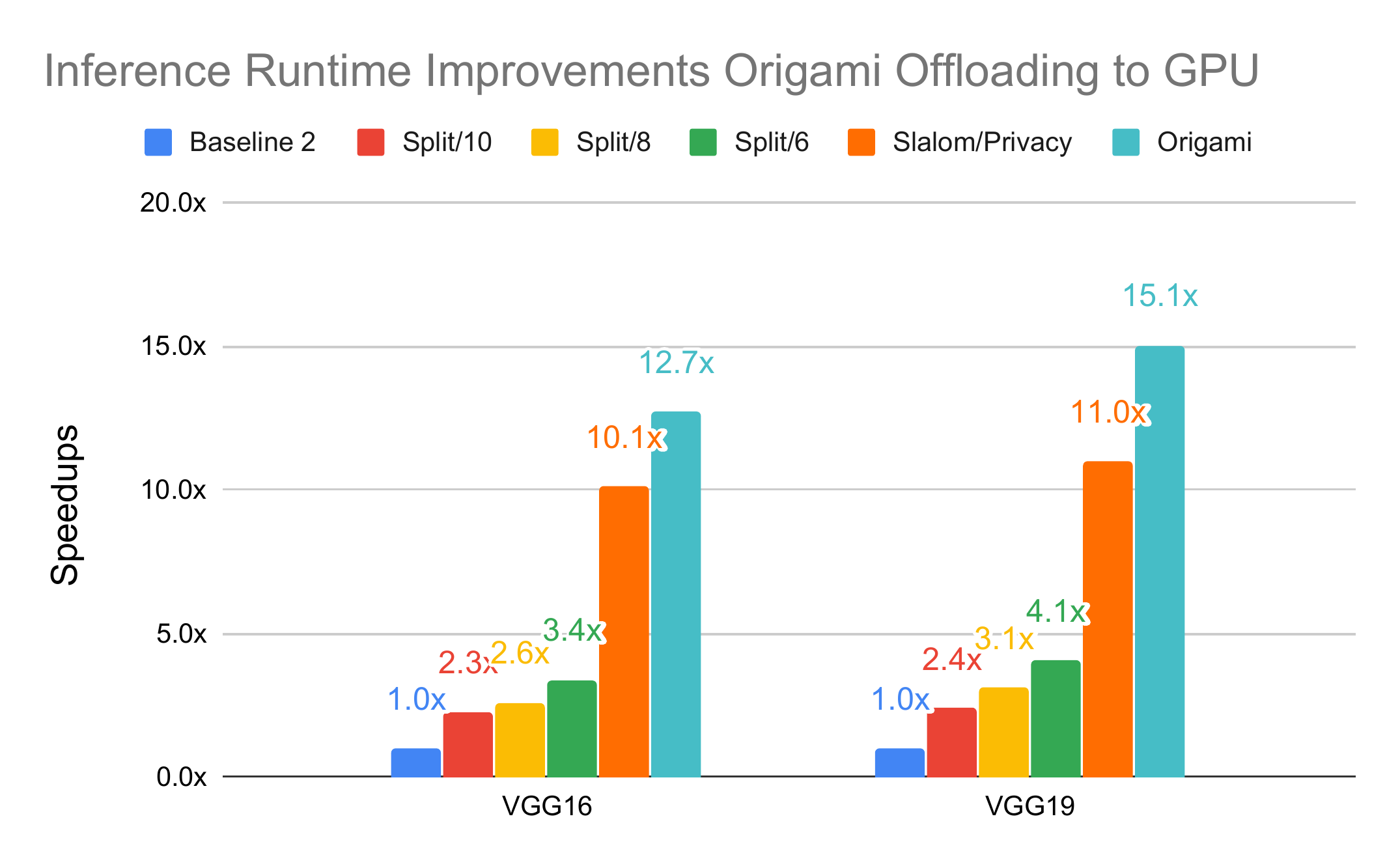}\\
  \caption{Inference Runtime offloading to GPU}
  \label{fig:IRGPU}
\end{figure}

Figure \ref{fig:IRGPU} shows the average runtime comparison of Origami inference with various baselines. In this figure all the offloaded computations are executed on GPU. Compared to executing the entire model within SGX (baseline 2), Slalom achieves 10x speedup on VGG16 (and 11x on VGG19). Origami achieves 12.7x and 15.1x speedups. Model splitting at layer 6 (Split/6), which is the minimum layer split needed to protect privacy, achieves only around 4X speedup, because the first six layers are essentially unable to take advantage of the vast GPU acceleration capability. 

\begin{figure}
\centering
  \includegraphics[width=8cm, height=6cm]{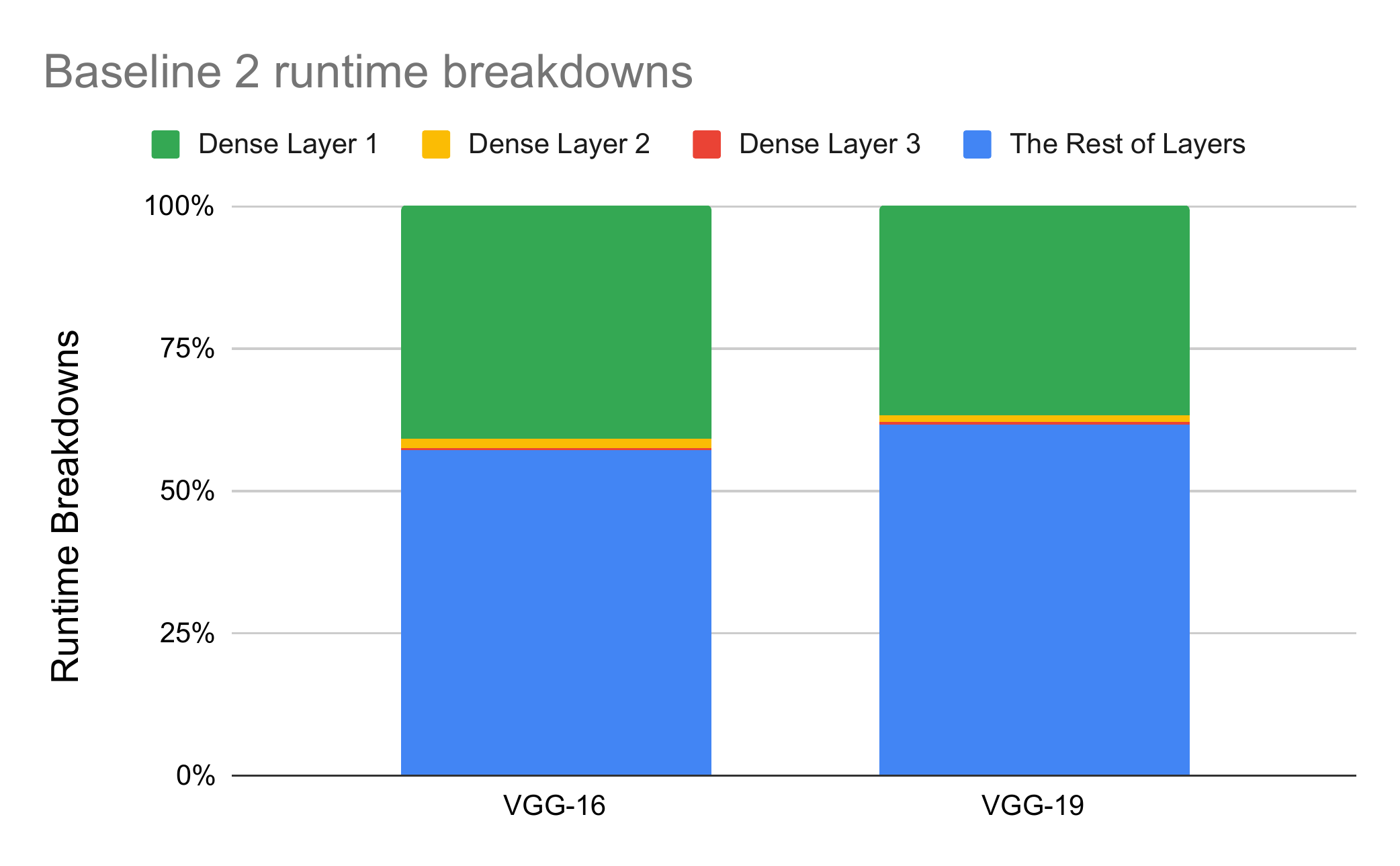}\\
  \caption{Baseline 2 runtime breakdown}
  \label{fig:bline2br}
\end{figure}

To understand the vast slowdown seen in the baseline, figure~\ref{fig:bline2br} presents runtime breakdown for baseline 2. The last three fully connected dense layers (Dense Layer 1,2 and 3) account for about 40\%  of baseline 2 runtime. We also analyzed how much time is spent in each layer doing the data processing and data movement in and out of SGX memory. We note that  around 50\% of the execution time in the fully connected dense layers is spent on data movement.  While Baseline 2's active memory footprint can fit within SGX physical protected memory, its dense layers are loaded on demand to prevent SGX memory limit overflow. Thus the baseline has to pay the penalty of fetching parts of parameters on demand. Note that pre-loading the entire model (our original baseline 1)  performs worse because it exceeds the SGX memory limit as well and pays significantly more data movement penalty. 

The bottleneck of Slalom/Privacy results from its need to blind and unblind intermediate features to guarantee privacy across each and every VGG layer. Processing intermediate features dominate Slalom/Privacy runtime. As mentioned earlier, unblinding or blinding 6MB features roughly takes 4 milliseconds and there are roughly 47MB and 51MB intermediates features to process in total for VGG-16 and VGG-19. Blinding and unblinding intermediate features takes roughly 62 and 68 milliseconds.

Origami achieves speedup by limiting the blinding overheads to just a few layers while enabling many layers to be offloaded to GPU.

\begin{figure}
\centering
  \includegraphics[width=8cm, height=6cm]{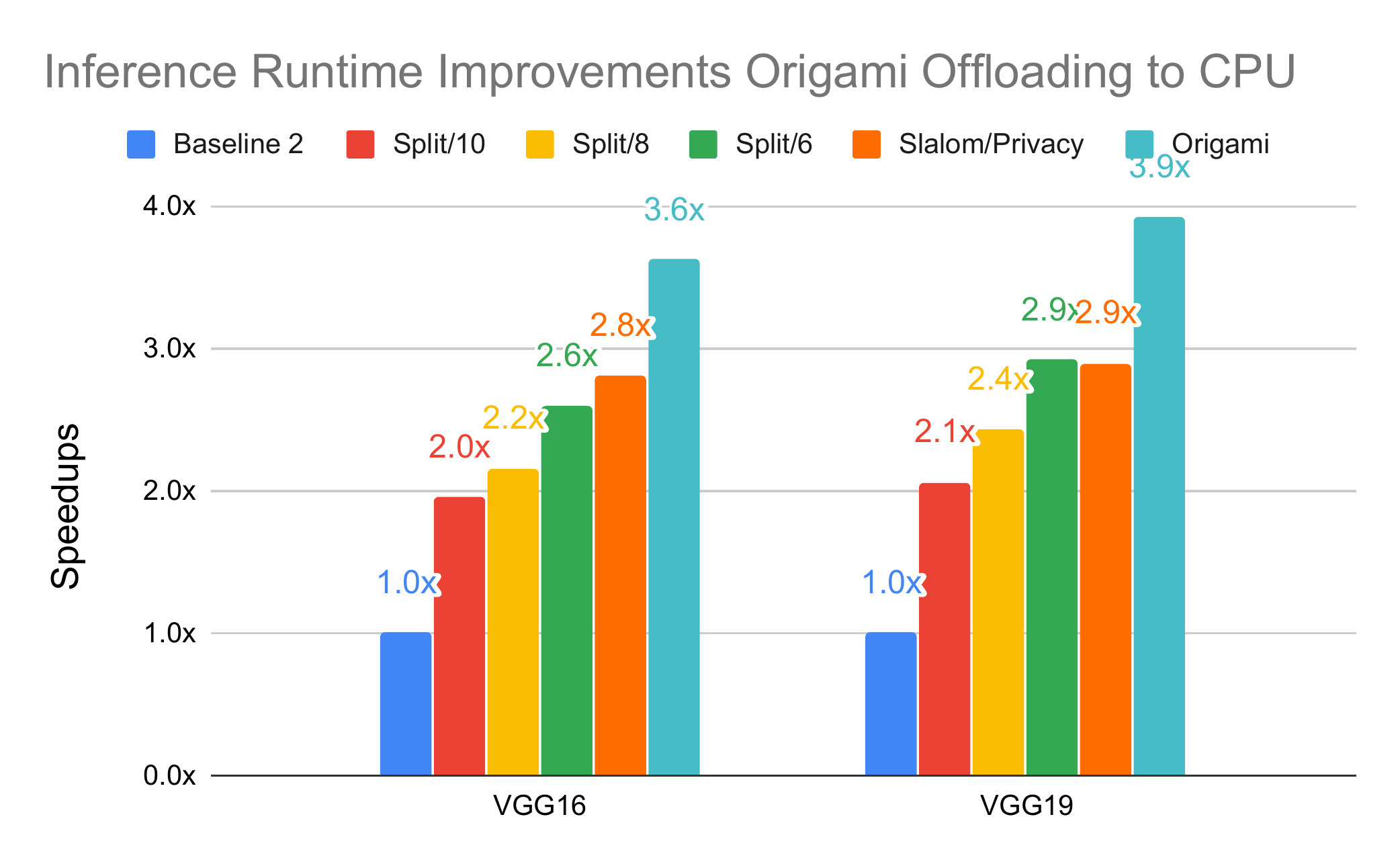}\\
  \caption{Inference Runtime offloading to CPU}
  \label{fig:IRCPU}
\end{figure}

Figure \ref{fig:IRCPU} shows the performance when all the offloaded computations are executed on CPU (no GPU usage). The highly parallel convolution operations were limited by the available CPU parallelism. Hence, Slalom achieves only about 2.9x speedup, while Origami achieves about 3.9x speedup compared with baseline 2 on VGG-19. Note that Slalom/Privacy has similar performance with Split/6 offloading to CPU. In this case the cost of blinding/unblindig is roughly the same as the computational overhead of executing the first six layers direcly on the SGX.  Slalom achieves smaller improvements by offloading linear layers when the intermediate feature map is too big and untrusted hardware is not significantly faster. 


\begin{table}[h!]
\centering
\caption{Enclave Memory Requirements for VGG16}
\begin{tabular}{||c | c||} 
 \hline
Model Type & Required Size \\
 \hline\hline
 Baseline 2 & 86MB\\
 \hline
 Split/6  & 29MB\\
 \hline
 Split/8  & 33MB\\
 \hline
 Split/10 & 35MB\\
 \hline
 Slalom/Privacy & 39MB\\
 \hline
  Origami & 39MB\\
  \hline
\end{tabular}
\label{table:1}
\end{table}

\subsubsection{Power Event Recovery}
After a power event, SGX enclave applications have to be reinitialized. Table \ref{table:2} shows the time to recover from a power event. Split models can recover much faster than  baseline 2 which need to reinitialize the model parameters before restarting the service. Since Split models need less memory inside SGX enclave, fewer pages are encrypted during enclave initialization, and less data are copied into SGX enclave during model creation. Origami and Slalom/Privacy will have similar recovery time because they have the same memory requirement.

\begin{table}[h!]
\centering
\caption{Recovery Time from Power Events for VGG16}
\begin{tabular}{||c | c||} 
 \hline
Model Type & Required Time \\
 \hline\hline
 Baseline 2 & 201ms\\
 \hline
 Split/6  & 51ms\\
 \hline
 Split/8  & 54ms\\
 \hline
 Split/10 & 59ms\\
 \hline
\end{tabular}
\label{table:2}
\end{table}

\subsubsection{Comparison with non-private inference}
\begin{figure}
\centering
  \includegraphics[width=8cm, height=6cm]{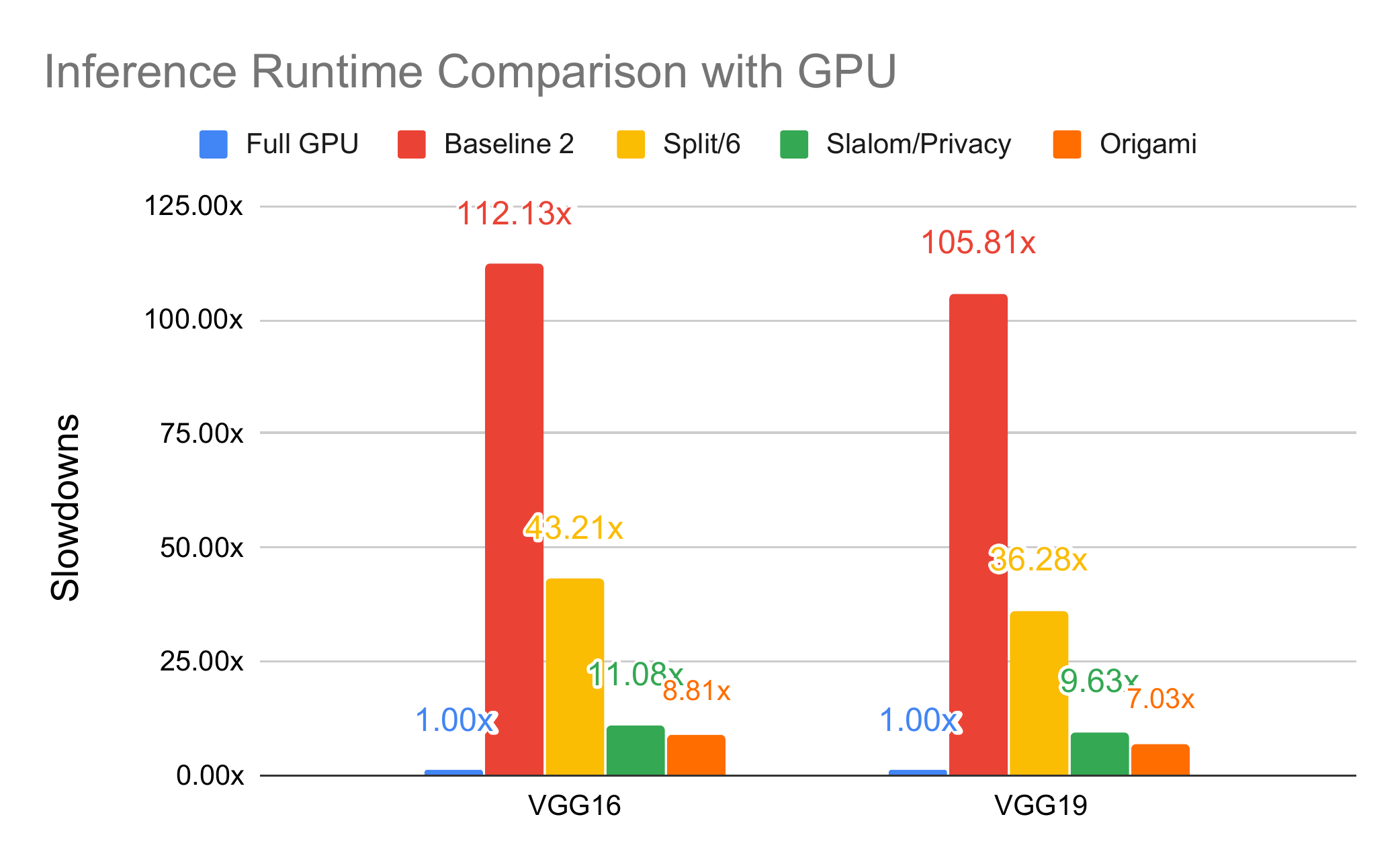}\\
  \caption{Inference runtime offloading to GPU}
  \label{fig:CPGPU}
\end{figure}

\begin{figure}
\centering
  \includegraphics[width=8cm, height=6cm]{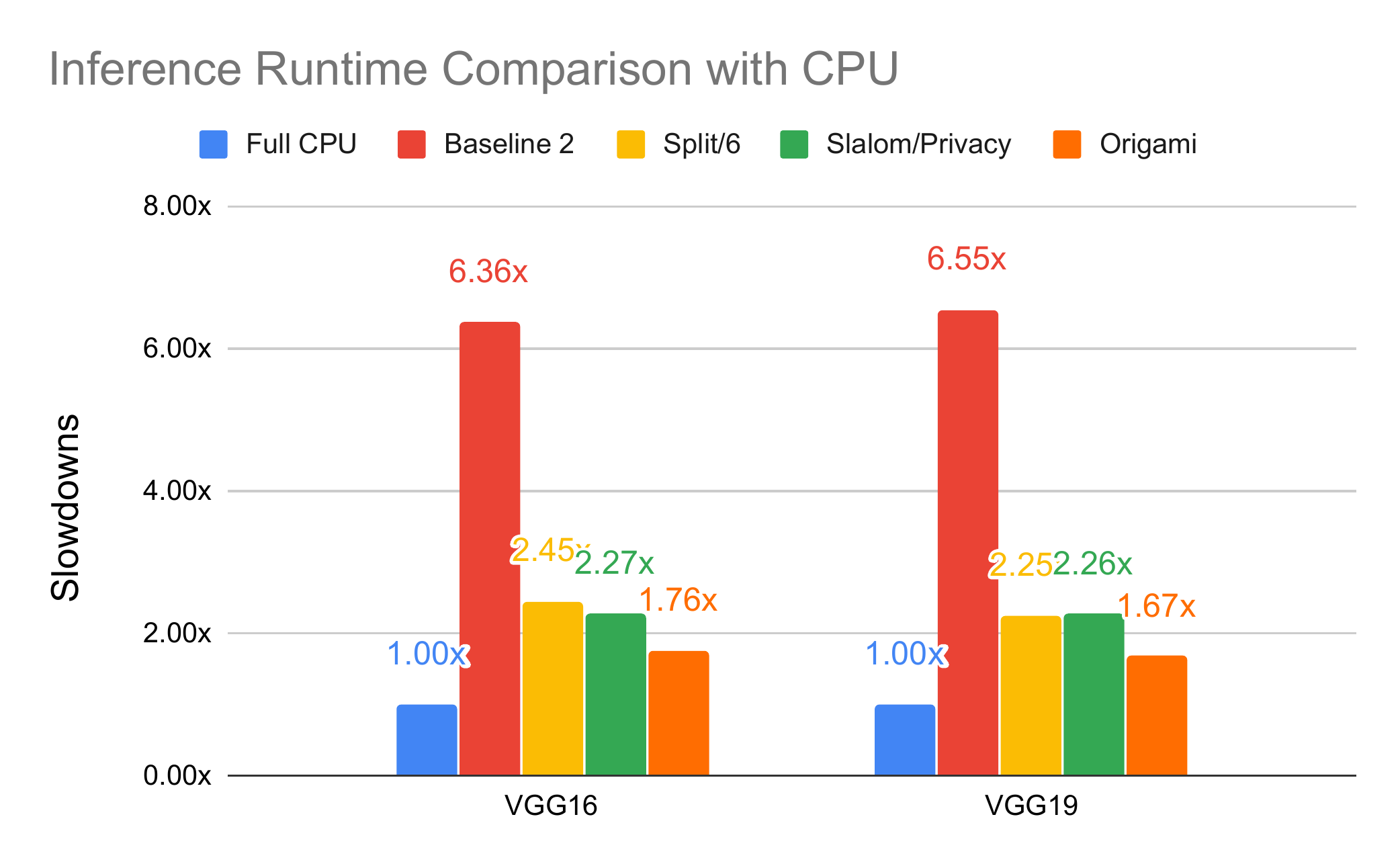}\\
  \caption{Inference runtime offloading to CPU}
  \label{fig:CPCPU}
\end{figure}

Figure \ref{fig:CPGPU} and figure \ref{fig:CPCPU} present relative inference runtime compared to a baseline where the entire model is executed in fast hardware without any privacy guarantees. Compared with CPU, Origami takes 1.7x longer at most. Origami takes about 8x longer when compared to running the entire model within a GPU.

The slowdown results from the use of SGX enclave. Origami and Slalom respectively pay about 47 and 62 milliseconds of penalty for SGX operations related to blinding, unblinding and non-linear operations for VGG-16. Since a GPU is well suited for highly parallel inference operations running the entire model within a GPU is still  faster, when no privacy guarantees are needed. These results show that supporting private execution environments within accelerators has significant usage potential in future. When compared with untrusted CPUs, Origami inference has reasonable performance overhead of 0.7x while providing strong privacy guarantees.

\subsection{Evaluation Summary}
When compared with the baselines, our experiments showed that Origami framework, while providing strong confidentiality guarantees, benefits from offloading computations to fast hardware. Offloading greatly reduces computation and memory requirements inside SGX enclave. Reduction in memory requirements enables Origami inference to recovery faster from unexpected power events. 
Reduced memory requirement for Origami also allows running more enclave applications simultaneously.
When compared with Slalom/Privacy, Origami inference benefits from reduced amount of computation assigned to the SGX enclave and has significantly lower inference latency.

\section{Related Work}
\label{sec:related_work}
Origami inference uses a combination of model partitioning, computational offloading and data blinding to provide fast and private inference. We discuss the related works for each technique separately.

\noindent \textbf{Model partitioning and offloading}: Previous efforts like \cite{Gordon12comet, odessa, maui, cloneCloud} focus on improving performance on mobile devices by partitioning and offloading computation. Neurosurgeon \cite{neurosurgeon}  partitions DNNs and onloads them to mobile devices to reduce latency and energy consumption. Although Origami performs model partitioning and offloading, it differs from all these works in it's focus on protecting privacy of the data being processed. Also, Origami only offloads computation to co-located CPUs and GPUs. Authors of \cite{DBLP:journals/corr/abs-1807-00969} propose a ternary DNN model partitioning approach to reduce the overhead of using enclave. Origami differs from this in multiple ways. First, Origami uses Slalom's blinding and unblinding and never performs linear operations inside the enclave. As a result, Origami has lower inference latency. Second, Origami protects privacy against a powerful c-GAN adversary\cite{DBLP:journals/corr/MirzaO14}.

\noindent \textbf{Image reconstruction}: A common adversarial objective on image classification networks is image reconstruction. Authors of work \cite{ImageInvertion} showed that intermediate feature maps from early layers in CNNs leak  information and that even non-adversarial models can be optimized to reconstruct input images based on these feature maps. Making such informative feature maps available for a powerful adversary can compromise privacy of input. In our approach we choose the partition point of CNN models such that the most informative feature maps are blinded inside SGX and only significant information curtailed feature maps are completely offloaded. We also choose to protect against reconstruction by evaluating against a conditional GAN adversary, since GAN based models are the current state-of-the-art adversaries. The potency of a GAN trained with conditional information has been demonstrated in works such as \cite{PacGAN, XGan}.

\noindent \textbf{Homomorphic encryption}: Another paradigm of trying to provide privacy guarantees when performing inference on machine learning models is the use of homomorphic encryption. Recent works \cite{homeomorphic, homeo2, homo2, homo3, gazelle} have explored using this technique to provide secure inference on models in the setting of adversaries trying to learn information about the data. Encryption and processing on encrypted data contribute to prohibitively large overheads and Origami inference is orders of magnitude faster than these methods. The only overhead Origami incurs is the cryptographic blinding and unblinding operations on the intermediate feature maps inside the enclave.

\noindent \textbf{Adding noise for privacy}: Authors of \cite{DBLP:journals/corr/abs-1809-03428} train auxiliary models to inject noise and perform data nullification before offloading data to the cloud for inference. They retrain the DNN used in the cloud whereas in Origami the DNN model is not modified. Shredder \cite{shredder} partitions the DNN models between the edge and the cloud, and adds learned noise to the intermediate data at the edge before offloading to the cloud. Shredder does not make use of the enclaves and performs computation on the edge device. Unlike Shredder where the noise is learned, in Origami the blinding factors are not learned.

\noindent \textbf{Side channel attacks}: Origami relies on the security guarantees provided by Intel SGX. However, SGX has been shown to be prone to side channel attacks \cite{sgxPectre, foreshadown} based on speculative execution bugs like Spectre and Meltdown \cite{spectre, meltdown}. Intel is making updates to it's hardware and SGX implementation to increase robustness against these attacks. Techniques proposed in \cite{stt, InvisiSpec} can be used to defend against bugs in speculative execution. Finally, Origami is a general framework  and can be applied to other enclave architectures like Sanctum \cite{sanctum}.


\section{Conclusion}
\label{sec:conclusion}
In cloud based inference services, protecting the privacy of the user data is very important. In this work, we proposed Origami Inference which leverages hardware enclaves to protect the privacy of the user data. With Origami inference we bring the performance of using an enclave close to the performance of executing an inference request outside of the enclave in a CPU or a GPU. Origami inference achieves this by combining model partitioning with the blinding of data inside the hardware enclaves. We demonstrate the privacy of Origami with a strong conditional GAN adversary and show excellent performance over strong baselines during evaluations. Moving forward, we plan to explore techniques to further reduce the amount of computation that need to take place inside the TEE.

\bibliographystyle{IEEEtranS}
\bibliography{0.privacy_main.bib}

\end{document}